\documentclass[preprint,3p]{elsarticle}

\usepackage{lineno,hyperref,enumitem}
\usepackage[usenames,dvipsnames,svgnames,table]{xcolor}
\usepackage{color,soul}
\usepackage{tcolorbox}
\usepackage{amssymb}
\usepackage{pifont}
\newcommand{\cmark}{\ding{51}}%
\newcommand{\xmark}{\ding{55}}%
\usepackage{makecell}
\usepackage{booktabs}
\usepackage{subcaption}

\newcolumntype{L}[1]{>{\raggedright\let\newline\\\arraybackslash\hspace{0pt}}m{#1}}
\newcolumntype{C}[1]{>{\centering\let\newline\\\arraybackslash\hspace{0pt}}m{#1}}
\newcolumntype{R}[1]{>{\raggedleft\let\newline\\\arraybackslash\hspace{0pt}}m{#1}} 

\xdefinecolor{gray95}{gray}{0.65}
\xdefinecolor{gray25}{gray}{0.1}

\usepackage{scalerel}

\modulolinenumbers[5]

\usepackage{tikz}
\newcommand*\circled[1]{\smash{\tikz[baseline=(char.base)]{
            \node[shape=circle,draw,inner sep=2pt] (char) {#1};}}}

\journal{Swarm and Evolutionary Computation}

\begin{document}

\begin{frontmatter}

\title{A Tutorial on the Design, Experimentation and Application of Metaheuristic Algorithms to Real-World Optimization Problems}

\author[tec]{Eneko Osaba\corref{cor1}}
\author[tec]{Esther Villar-Rodriguez}
\author[tec,upv]{Javier~Del Ser}
\author[mal]{\\Antonio J. Nebro}
\author[ugr]{Daniel Molina}
\author[upm]{Antonio LaTorre}
\author[ntu]{\\Ponnuthurai N. Suganthan}
\author[civ]{Carlos A. Coello Coello}
\author[ugr]{Francisco Herrera}

\cortext[cor1]{Corresponding author. TECNALIA, Basque Research and Technology Alliance (BRTA), 48160 Derio, Spain. Phone: +34 677 521 434. Email address: eneko.osaba@tecnalia.com}
\address[tec]{TECNALIA, Basque Research and Technology Alliance (BRTA), 48160 Derio, Spain}
\address[upv]{University of the Basque Country (UPV/EHU), 48013 Bilbao, Spain}
\address[ugr]{DaSCI Andalusian Institute of Data Science and Computational Intelligence, University of Granada, 18071 Granada, Spain}
\address[mal]{Dept. de Lenguajes y Ciencias de la Computaci\'on, ITIS Software, Universidad de M\'alaga, Spain}
\address[civ]{CINVESTAV-IPN, 07360 Mexico, DF, Mexico}
\address[ntu]{Nanyang Technological University, 639798, Singapore}
\address[upm]{Center for Computational Simulation, Universidad Polit\'ecnica de Madrid, Madrid, Spain}

\begin{abstract}
In the last few years, the formulation of real-world optimization problems and their efficient solution via metaheuristic algorithms has been a catalyst for a myriad of research studies. In spite of decades of historical advancements on the design and use of metaheuristics, large difficulties still remain in regards to the understandability, algorithmic design uprightness, and performance verifiability of new technical achievements. A clear example stems from the scarce replicability of works dealing with metaheuristics used for optimization, which is often infeasible due to ambiguity and lack of detail in the presentation of the methods to be reproduced. Additionally, in many cases, there is a questionable statistical significance of their reported results. This work aims at providing the audience with a proposal of good practices which should be embraced when conducting studies about metaheuristics methods used for optimization in order to provide scientific rigor, value and transparency. To this end, we introduce a step by step methodology covering every research phase that should be followed when addressing this scientific field. Specifically, frequently overlooked yet crucial aspects and useful recommendations will be discussed in regards to the formulation of the problem, solution encoding, implementation of search operators, evaluation metrics, design of experiments, and considerations for real-world performance, among others. Finally, we will outline important considerations, challenges, and research directions for the success of newly developed optimization metaheuristics in their deployment and operation over real-world application environments.
\end{abstract}


\end{frontmatter}

\section{Introduction} 
\label{sec:intro}

The formulation and solution of optimization problems through the use of metaheuristics has gained an increasing popularity over the last decades within the Artificial Intelligence community \cite{hussain2019metaheuristic,boussaid2013survey}. This momentum has been propelled by the emergence and progressive maturity of new paradigms related to problem modeling (e.g., large scale optimization, transfer optimization), as well as by the vibrant activity achieved in the Swarm Intelligence and Evolutionary Computation fields \cite{kennedy2006swarm,eiben2015evolutionary,del2019bio}. In this regard, there are several crucial aspects and phases that define a high-quality research work within these specific areas. Each of these aspects deserves a painstaking attention for reaching the always desirable replicability and algorithmic understanding. Moreover, these efforts should be intensified if the conducted research has the goal of being deployed in real-world scenarios or applications.

This last aspect unveils one of the most backbreaking challenges that researchers face these days. It is relatively easy to find in the literature really meaningful studies around theoretical and synthetic applications of optimization problems and their solution using metaheuristic algorithms \cite{yang2018mathematical,pranzo2016iterated,vidal2015hybrid}. However, it is less frequent to find thoughtful and comprehensive studies focused on real-world deployments of optimization systems and applications. 

Besides that, the challenge is even more arduous when the main goal of the research is to put in practice a previously published theoretical and experimental study. There are two main reasons that generate this complicated situation. First, it is difficult to find studies that allow the work carried out to be transferred to practical environments without requiring a significant previous effort. On the other hand, the second reason points to the lack of a practical guide for helping researchers to outline all the steps that a research work should meet for being reproducible such that it contemplates both theoretical and real-world deployment aspects.

With this in mind, it is appropriate to claim that the gap between theoretical and real-world oriented research is still evident in the research on metaheuristics for optimization that is being conducted nowadays. This gap is precisely the main motivation for the present study in which we propose a methodology of design, development, experimentation, and final deployment of metaheuristic algorithms oriented to the solution of real-world problems. The guidelines provided here will tackle different pivotal aspects that a metaheuristic solver should efficiently address for enhancing its replicability and to facilitate its practical application.

To rigorously meet the objective proposed in this paper, each of the phases that define a high-quality research are analyzed. This analysis is conducted from a critical but constructive approach towards amending misconceptions and bad methodological habits, with the aim of ultimately achieving valuable research of practical utility. To this end, the analysis made for each phase incorporates a prescription of application-agnostic guidelines and recommendations that should be followed by the community to foster actionable metaheuristic algorithms, namely, metaheuristic methods designed and tested in a principled way, with a view towards ensuring their actual use in real-world applications.

Over the years, several efforts have been made by renowned researchers for establishing firm foundations that guide practitioners to conduct rigorous research studies \cite{danka2013statistically,kendall2016good,jaszkiewicz2004evaluation,chiarandini2007experiments}. All these previous studies have significantly contributed to the standardization of some important concepts. However, the majority of these works are focused on some specific steps or phases of the whole optimization process, while others focus on very specific knowledge domains. These remarkable studies will be analyzed in upcoming sections. Such works certainly helped us to highlight the main novelty of the methodology proposed here, which is the deeming of each step that makes up a real-world oriented optimization research. We cover from the early phase of problem modeling to the validation and practical operation of the developed algorithm. Therefore, the main contributions and analysis of this tutorial are focused on the following issues:

\begin{itemize}[leftmargin=*]
    \item \textit{Problem Modeling and Mathematical Formulation}: this first step is devoted to the modeling and mathematical formulation of the optimization problem, which is guided by a previously conducted conceptualization. 
    
    \item \textit{Algorithmic Design, Solution Encoding and Search Operators}: this phase is entirely dedicated to the design and the implementation of the metaheuristic algorithm.
    
    \item \textit{Performance Assessment, Comparison and Replicability}: this is a crucial step within the optimization problem solving process, and it is devoted to the correct evaluation of the algorithms developed, and to the replicability and consistency of the research. 
    
    \item \textit{Algorithmic Deployment for Real-World Applications}: once the metaheuristic is developed and properly tested, this last phase is dedicated to the deployment of the method in a real environment.

\end{itemize}

The remainder of the paper is structured as follows. In Section~\ref{sec:meta}, the history of problem solving through metaheuristics is briefly outlined, underscoring the related inherent methodological uncertainties. In 
Section~\ref{sec:workflow}, we introduce a reference workflow which will guide the whole methodology. We also highlight in this section some of the most important related works already published in the literature and their connection with the methodology proposed in the present paper. Our proposed practical procedure is described in detail in Sections~\ref{sec:metho},~\ref{sec:design},~\ref{sec:assesment}, and~\ref{sec:deployment}. Additionally, a summary of good practices at each specific phase of the complete problem solving process is provided in 
Section~\ref{sec:recommendations}. The tutorial ends with a discussion on future research lines of interest for the scope of this tutorial, 
Section~\ref{sec:prospects}, followed by our concluding remarks provided in Section~\ref{sec:conc}.

\section{Problem solving using Metaheuristics: a Long History with Methodological Uncertainties}
\label{sec:meta}

Optimization problems and their efficient handling has received extensive attention throughout the years. The appropriate solution of extraordinarily complex problems usually entails the use of significant computation resources \cite{hochba1997approximation,papadimitriou1998combinatorial,papadimitriou1991optimization}. This computational complexity, along with their ease of application to real-world situations, has made of the optimization field one of the most intensively studied by the current artificial intelligence community. This scientific interest has led to the proposal of a plethora of solution approaches by a considerable number of researchers and practitioners. Arguably, the most successful methods can be grouped into three different categories: (1) exact methods, (2) heuristics, and (3) metaheuristics. As stated previously, this study will sharpen its focus on the last of these categories. 

Metaheuristics can be divided into different categories depending on their working philosophy and inspiration \cite{blum2003metaheuristics,dokeroglu2019survey}. For a better understanding of the situation described in this paper, it is interesting to put emphasis on a specific branch of knowledge related to metaheuristics and optimization problem solving: bio-inspired computation \cite{yang2013swarm}. In the last two decades, a myriad of bio-inspired approaches have been applied to different problems, some of which have shown remarkable performance. This growing attention has led to an extraordinary increase in the amount of relevant published material, usually focused on the adaptation, improvement, and analysis of a variety of methods that have been previously reported in the specialized literature.

Several reasons have contributed to this situation. Probably, the most important cornerstone was the birth of the branches which are known today as {\em Evolutionary Computation} and {\em Swarm Intelligence} \cite{bonabeau1999swarm,de2006evolutionary}. The main representative techniques within these streams are the genetic algorithm (GA, \cite{genetic1,genetic2}), particle swarm optimization (PSO, \cite{PSO}), and ant colony optimization (ACO, \cite{dorigo1997ant}). Being more specific, it was PSO, thanks to its overwhelming success and novelty, the one that decisively influenced the creation of a plethora of bio-inspired methods, which clearly inherit its main philosophy \cite{molina2020comprehensive}. 

In spite of the existence of an ample collection of classical and sophisticated solvers proposed in both past and recent literature, an important segment of the research community continues scrutinizing the natural world seeking to formulate new metaheuristics that mimick new biological phenomena. This fact has entailed the seeding of three different problems in the community, which are now deeply entrenched. We list these problems below:

\begin{itemize}[leftmargin=*]
    \item Usually, the proposed novel methods are not only unable to offer a step forward for the community, but also augment the skepticism of critical researchers. These practitioners are continuously questioning the need for new methods, which apparently are very similar to previously published ones. Some studies that have discussed this problem are \cite{sorensen2015metaheuristics}, \cite{sorensen2018history} or \cite{del2019bio}.
    \item The uncontrolled development of metaheuristics contributes to grow an already overcrowded literature, which is prone to generate ambiguities and insufficiently detailed research contributions. This uncontrolled growth is splashing the research community with a large number of articles whose contents is not replicable and in some cases, it may be even unreliable. The reason is the ambiguity and lack of detail in the presentation of the methods to be replicated and the questionable statistical significance of their reported results.
    \item Most of the proposed methods are tested over synthetic datasets and generally compared with classical and/or representative metaheuristics. This fact also involves the generation of two disadvantages. First of all, the sole comparison with classical techniques has led to unreliable and questionable findings. Second, the approaches proposed in these publications is usually difficult to deploy in real-world environments, requiring huge amounts of time and effort to make them work. Finally, being aware of the rich related literature currently available, today's scientific community must turn towards the proposal of practical and real-world applications of metaheuristic algorithms. This goal cannot be reached if part of the community continues delving into the proposal of new solution schemes which, in most cases, don't seem to be fully justified. 
\end{itemize}

For reversing this non-desirable situation, we provide in this work a set of good practices for the design, experimentation, and application of metaheuristic algorithms to real-world optimization problems. Our main goals with the methodology proposed in this paper is to guide researchers to conduct fair, accurate, and shareable applied studies, deeming all the spectrum of steps and phases from the inception of the research idea to the final real-world deployment. 

As has been pointed out in the introduction, some dedicated efforts have been conducted before with similar purposes. Some of these papers are currently cornerstones for the community, guiding and inspiring the development of many high-quality studies. In \cite{derrac2011practical}, for instance, a tutorial on the use of non-parametric statistical tests for the comparison of evolutionary and swarm intelligence metaheuristics is presented. In that paper, some essential non-parametric procedures for conducting both pairwise and multiple comparisons are detailed and surveyed. A similar research is introduced in \cite{danka2013statistically}, in which a procedure for statistically comparing heuristics is presented. The goal of that paper is to introduce a methodology to carry out a statistically correct and bias-free analysis.

In \cite{braysy2005vehicle}, a detailed study on the Vehicle Routing Problem with Time Windows is presented, in which several guides are offered for the proper design of solutions and operators, among other remarkable aspects. In any case, one of the most valuable parts of this research is the in-depth discussion on how heuristic and metaheuristic methods should be assessed and compared. An additional interesting paper is \cite{osaba2018good}, which proposes a procedure to introduce new techniques and their results in the field of routing problems and combinatorial optimization problems. Furthermore, in a previously cited paper, Sorensen \cite{sorensen2015metaheuristics} also provides some good research practices to follow in the implementation of novel algorithms.

The difficulty of finding standards in optimization research in terms of significant laboratory practices is the main focus of the work proposed in \cite{kendall2016good}. Thus, the authors of that work suggest some valuable recommendations for properly conducting rigorous and replicable experiments. A similar research is proposed in the technical report published by Chiaraindini et al. \cite{chiarandini2007experiments}. In that report, the authors formalize several scenarios for the assessment of metaheuristics through laboratory tests. More specific is the study presented in \cite{eggensperger2019pitfalls}, focused on highlighting the many pitfalls in algorithm configuration and on introducing a unified interface for efficient parameter tuning.

It is also interesting to mention the work proposed in \cite{eiben2002critical}, which introduces some good practices in experimental research within evolutionary computation. Focused also in evolutionary computation, the authors of \cite{vcrepinvsek2014replication} highlight some of the most common pitfalls researchers make when performing computational experiments in this field, and they provide a set of guidelines for properly conducting replicable and sound computational tests. A similar effort is made in \cite{latorre2020fairness} but focused on bio-inspired optimization. The literature contemplates additional works of this sort, such as \cite{hansen2016coco}.

The methodologies mentioned up to now revolve around two key aspects in optimization: efficient algorithmic development and rigorous assessment of techniques. In addition to that, it is also possible to find in the literature good practices about the modeling and formulation of the optimization problem itself. This issue is equally important to the others that have been previously mentioned, and not dealing properly with it, usually becomes a source of multiple uncertainties and inefficiencies. In~\cite{edmonds_2008}, for example, Edmonds provides a complete guide for properly formulating mathematical optimization problems. The author of that paper highlights the importance of analyzing the complexity of problems, which is crucial for choosing and justifying the use of a solution method. He also stresses the importance of carefully defining three different ingredients that make up an optimization problem: instances, solutions, and costs.  

Also related are the works conducted in \cite{jaszkiewicz2004evaluation} and \cite{huang2007problem}, both dedicated to multi-objective problems. Moreover, in its successful book \cite{kumar2019research}, Kumar dedicates a complete section to guide researchers in the proper definition of optimization problems. This book is especially valuable for newcomers in the area due to its informative nature. Apart from these generic approaches, valuable works of this sort can be found in the literature devoted to some specific knowledge domains, such as the ones presented in \cite{jie2019two} and \cite{delorme2016bin}.

As indicated before, the community has made remarkable efforts to establish some primary lines which should guide the development of high-quality, transparent, and replicable research. The main original contribution of the methodology proposed in this paper is the consideration of the full procedure related to a real-world oriented optimization research, covering from the problem modelling to the validation and practical operation of the developed systems. Finally, Table~\ref{tab:summary} summarizes the state of the art outlined in this section. We also depict the main contribution of our proposal in comparison with each of the works described there. 

\begin{table}[ht!]
    \footnotesize
    \centering
    \resizebox{\columnwidth}{!}{
        \begin{tabular}{cC{0.12\textwidth}C{0.14\textwidth}C{0.12\textwidth}C{0.12\textwidth}L{0.35\textwidth} }
        \toprule
            & {\scriptsize Problem Modeling and Mathematical Formulation} & {\scriptsize Algorithmic Design, Solution Encoding and Search Operators} & \scriptsize Performance Assessment, Comparison and Replicability & \scriptsize Algorithmic Deployment for Real-World Applications & Focus\\ \midrule
            
            \cite{sorensen2015metaheuristics} & \centering \xmark  & \centering \cmark & \centering \xmark & \centering \xmark  & \scriptsize Bio-inspired and swarm intelligence algorithms \\ 
            
            \cite{danka2013statistically} & \centering \xmark  & \centering \xmark & \centering \cmark & \centering \xmark  &  \scriptsize Resource-constrained project scheduling problems \\ 
        
            \cite{kendall2016good} &  \centering \xmark  & \centering \xmark & \centering \cmark & \centering \xmark  & ---\\ 
            
            \cite{jaszkiewicz2004evaluation} &  \centering \cmark  & \centering \xmark & \centering \cmark & \centering \xmark  & \scriptsize Multi-objective problems\\
            
            \cite{chiarandini2007experiments} & \centering \xmark  & \centering \xmark & \centering \cmark & \centering \xmark  & --- \\ 
            
            \cite{derrac2011practical} & \centering \xmark  & \centering \cmark & \centering \cmark & \centering \xmark & \scriptsize Swarm Intelligence \\ 
            
            \cite{braysy2005vehicle} & \centering \xmark  & \centering \cmark & \centering \cmark & \centering \xmark & \scriptsize Vehicle Routing Problems with Time Windows\\
            
            \cite{osaba2018good} &  \centering \xmark  & \centering \centering \cmark & \centering \cmark & \centering \xmark  & \scriptsize Vehicle Routing Problems \\ 
            
            \cite{eggensperger2019pitfalls} &  \centering \cmark  & \centering \xmark & \centering \cmark & \centering \xmark & Common pitfalls and good practices for parameter tuning\\
            
            \cite{eiben2002critical} &  \centering \cmark  & \centering \xmark & \centering \cmark & \centering \xmark   & --- \\
            
            \cite{vcrepinvsek2014replication} &  \centering \cmark  & \centering \xmark & \centering \cmark & \centering \xmark & --- \\
            
            \cite{latorre2020fairness} & \centering \xmark  & \centering \xmark & \centering \cmark & \centering \xmark & \scriptsize Bio-inspired Optimization \\
            
            \cite{hansen2016coco} &  \centering \cmark  & \centering \xmark & \centering \cmark & \centering \xmark & Numerical optimization algorithms in a black-box scenarios\\
            
            \cite{edmonds_2008} &  \centering \cmark  & \centering \xmark & \centering \xmark & \centering \xmark  & --- \\ 
            
            \cite{huang2007problem} &  \centering \cmark  & \centering \xmark & \centering \cmark & \centering \xmark   & \scriptsize Multi-objective problems \\ 
            
            \cite{kumar2019research} &   \centering \cmark  & \centering \xmark & \centering \xmark & \centering \xmark   & \scriptsize Newcomers \\ 
            
            \cite{jie2019two} &  \centering \cmark  & \centering \xmark & \centering \xmark & \centering \xmark  & \scriptsize Two-echelon Capacitated Vehicle Routing Problem\\ 
            
            \cite{delorme2016bin} & \centering \cmark  & \centering \cmark & \centering \xmark & \centering \xmark  & \scriptsize Bin Packing Problems\\ 
            
            \midrule
            
            This work & \centering \cmark  & \centering \cmark & \centering \cmark & \centering \cmark   &  --- \\
            \bottomrule
        \end{tabular}
    }
        \caption{Summary of the literature review, and comparison with our proposed methodology.}
    \label{tab:summary}
\end{table}

\section{Solving Optimization Problems with Metaheuristic Algorithms: a Reference Workflow}
\label{sec:workflow}

\begin{figure}[t]
\centering
\includegraphics[width=0.95\hsize]{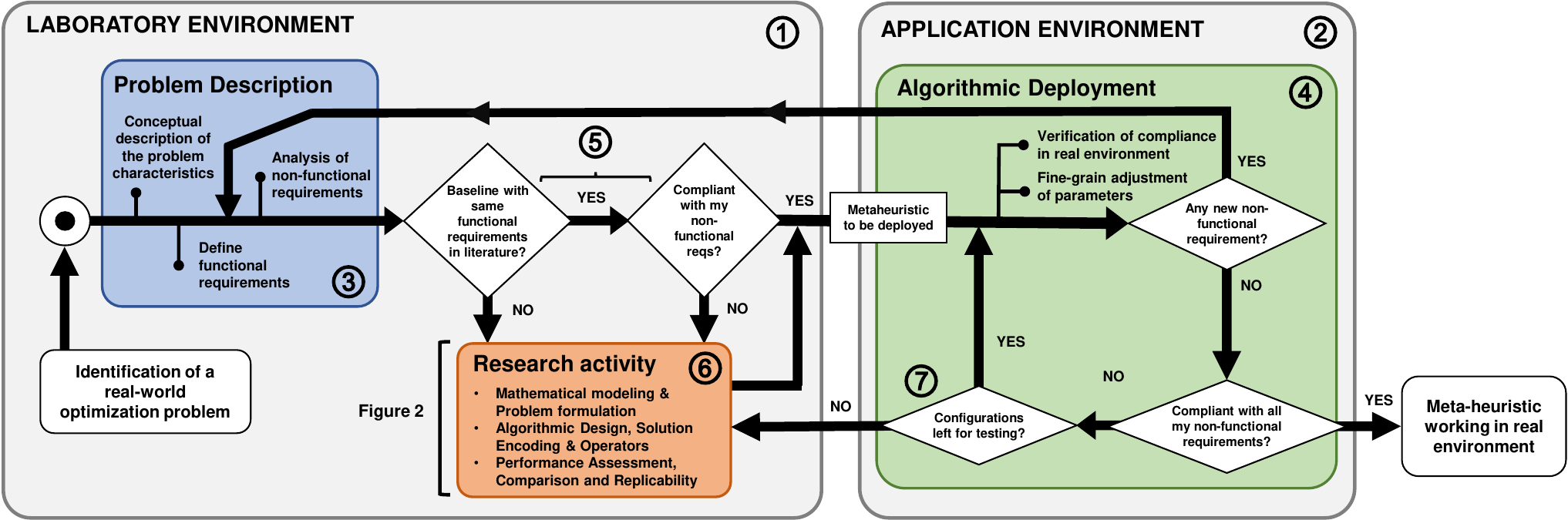}
\caption{Phase 1 of the reference workflow for solving optimization problems with metaheuristic algorithms.}
\label{fig:workflow1}
\end{figure}

\begin{figure}[t]
\centering
\includegraphics[width=0.9\hsize]{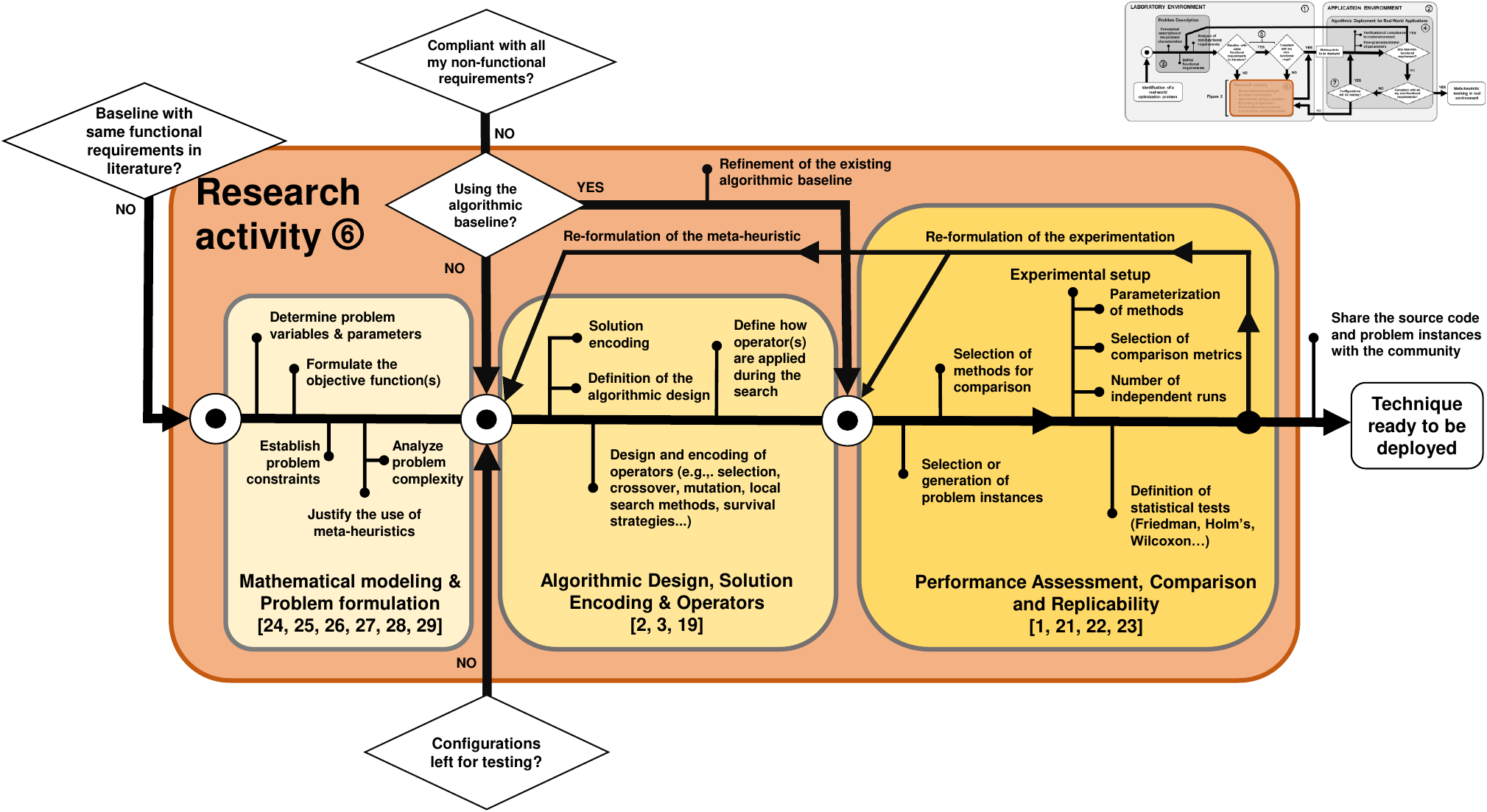}
\caption{Phase 2 of the reference workflow for solving optimization problems with metaheuristic algorithms.}
\label{fig:workflow2}
\end{figure}

In this section, we introduce the reference workflow that describes our methodological proposal. Our main intention is to establish this procedure as a reference, considering its adoption a must for properly conducting both theoretical and practical rigorous, thorough, and significant studies related to metaheuristic optimization. Thus, Figure~\ref{fig:workflow1} and Figure~\ref{fig:workflow2} represent this reference workflow, which will serve as a guide for the remaining sections of this paper.

Thus, we have used two different high-level schemes to describe our methodology graphically. The first one (Figure~\ref{fig:workflow1}) is conceived as the general scheme, and it contemplates the problem description \circled{3}, analysis, and development of the selected solution approach (5-6), and the deployment of the solution \circled{4}. On the other hand, the second scheme (Figure~\ref{fig:workflow2}) is completely devoted purely to the research activity (stage \circled{6} in Figure~\ref{fig:workflow1}). In another short glimpse, we can also see how we have devised two different development environments. Specifically, problem description, baseline analysis, and research activity are conducted in a laboratory environment \circled{1}, while the algorithmic deployment is conducted in an application environment \circled{2}.

Focusing our attention on the first workflow, the whole activity starts with the existence of a real problem that should be efficiently tackled. The detection of this problem and the necessity of addressing it, triggers the beginning of the research, whose first steps are the conceptual definition of the problem and the definition and analysis of both functional and non-functional requirements \circled{3}. It should be clarified here that this first description of the problem is made at a high-level, focusing on purely conceptual issues. Due to the nature of this first step, the presence of final stakeholders is highly recommended in addition to researchers and developers.

Regarding functional requirements, it is hard to find a canonical definition \cite{glinz2007non}, but they can be referred to as \textit{what the product must do} \cite{robertson2012mastering} or \textit{what the system should do} \cite{sommerville2001software}. Furthermore, the establishment of functional requirements involves the definition of the objective (or objectives in case of multi-objective problems) function to be optimized and the equality and inequality constraints (in case of dealing with a problem with side constraints). On the other hand, there is no such consensus for non-functional requirements. Davis defines them as \textit{the required of overall attributes of the system, including portability, reliability, efficiency, human engineering, testability, understandability and modificability} \cite{davis1993software}. Robertson and Robertson describe them as \textit{a property, or quality, that the product must have, such as an appearance, or a speed or accuracy properties} \cite{robertson2012mastering}. More definitions can be found in \cite{glinz2007non}. In any case, these objectives are crucial for the proper election of the solution approach, and the non-consideration of them can lead to the re-design of the whole research, involving both economical and time costs. This paramount importance is the reason why, in this work, we put special attention on highlighting the impact of the consideration or non-consideration of these non-functional objectives (of a fair and comprehensive description of the non-functional requirements). In fact, many of the research contributions available in the literature are focused on the pure fulfillment of functional requisites, making them hard to be properly deployed in the real world. Thus, we can see the meeting of non-functional objectives as the key for efficiently transitioning from the laboratory \circled{1} to the application environment \circled{2}. 

After this first conceptual phase, it is necessary to scrutinize the related literature and scientific community for finding an appropriate baseline \circled{5}. The main objective of this process is to find a public shared library or baseline that fits with the previously fixed functional requirements. In the positive case, the next step is to analyze whether these findings are theoretically compliant with all the outlined non-functional requirements. The published research activity is usually carried out under trivial and unofficial laboratory specifications with a short-sighted design mostly concentrated on the ``what'' (functional objectives) but not on the feasibility of ``the how''. The recommended good practice is filtering out research that has allegedly gone through from the lab hypothesis to the demanding real-world conditions. On the contrary, when assuming that the baseline does not satisfy or reckon these non-functional requirements, the research activity will first include procedures to evaluate the baseline viability so as to decide whether the baseline is still a potential workaround or has to be discarded \circled{6}. Finally, if both actions are positively solved, the investigation is considered ready to go through the deployment phase \circled{4}.

At this point, it is important to highlight that the so-called \textit{Algorithmic Deployment for Real-World Application} phase \circled{4} (detailed in Section~\ref{sec:deployment}), considered as a cornerstone in our methodology, can receive as input an algorithm directly drawn from a public library \circled{5}, or a method developed ad-hoc as a result of a thorough research procedure \circled{6}. At this phase, it could be possible to face the emergence of new non-functional objectives, implying the re-analysis of the problem (going back to \circled{3}) for the sake of deeming all the newly generated necessities. 

On the contrary, if all the non-functional requirements are considered but not fully met, further re-adjustments are necessary. In this scenario, additional minor adaptations should be made over the metaheuristic if further configurations are left to test \circled{7}. Nevertheless, if these minor adjustments do not result in a desirable performance of the algorithm, the process should re-iterate starting from the \textit{Algorithmic Design, Solution Encoding and Search Operators} phase (part of Workflow 2, Figure~\ref{fig:workflow2}, and detailed in Section~\ref{sec:design}), which may involve a re-design and re-implementation of (or even a new) our metaheuristic solution \circled{6}. Finally, if none of the above deviations occur and the performance of the metaheuristic meets the initially established objectives, the problem can be considered solved and the research completely finished after the final deployment of the algorithm in a real environment.

In another vein, Figure~\ref{fig:workflow2} depicts the second part of our workflow, which is devoted to the work related to research development. As can be easily seen in this graphic, this workflow has three different entry points, depending on the status of the whole activity. Furthermore, this phase is divided into three different and equally important sequential stages. These phases and how they are reached along the development process are detailed next:

\begin{itemize}[leftmargin=*]
    \item \textit{Problem Modeling and Mathematical Formulation} (Section~\ref{sec:metho}): This first step should be entirely devoted to the modeling and mathematical formulation of the optimization problem, which should be guided by the previously conducted conceptualization. The entry to this part of the research should be materialized if the problem to solve has not been tackled in the literature before, or in case of the non-existence of an adapted baseline or library. 
    
    \item \textit{Algorithmic Design, Solution Encoding and Search Operators} (Section~\ref{sec:design}): This second stage should be devoted to the design and implementation of the metaheuristic method. It should also be highlighted that another research branch could also be conducted, which is the refinement of a baseline or library already found in the scientific community.
    
    \item \textit{Performance Assessment, Comparison and Replicability} (Section~\ref{sec:assesment}): Once the algorithmic approach is developed (or refined), the performance analysis of the technique should be carried out. This is a crucial phase within the optimization problem solving process, and the replicability and consistency of the research clearly depend on the good conduction of this step. Furthermore, once the quality of the algorithms has been tested over the theoretical problem, it should be deployed in a real environment (\textit{Algorithmic Deployment for Real-World Application} phase, Figure~\ref{fig:workflow1}).

\end{itemize}

Once we have introduced and described our envisioned reference workflow, we outline in the following sections all the good practices that researchers and practitioners should follow for conducting high-quality, real-world oriented research.

\section{Problem Modeling and Mathematical Formulation} 
\label{sec:metho}

Once the analyst and domain expert have agreed upon the conceptual definition and the requirements to be met by the solution, the research activity gets started. All these inputs (conceptual description and functional/non-functional requirements) will be tracked along the whole workflow and be more approachable depending on the specific stage. At the problem modeling phase, the ``what'' contained in the conceptualization and functional requirements have to be perfectly clear and comprehensive enough to be fairly translated into a mathematical formulation.
\begin{figure}[h]
\centering
\includegraphics[width=1.0\hsize]{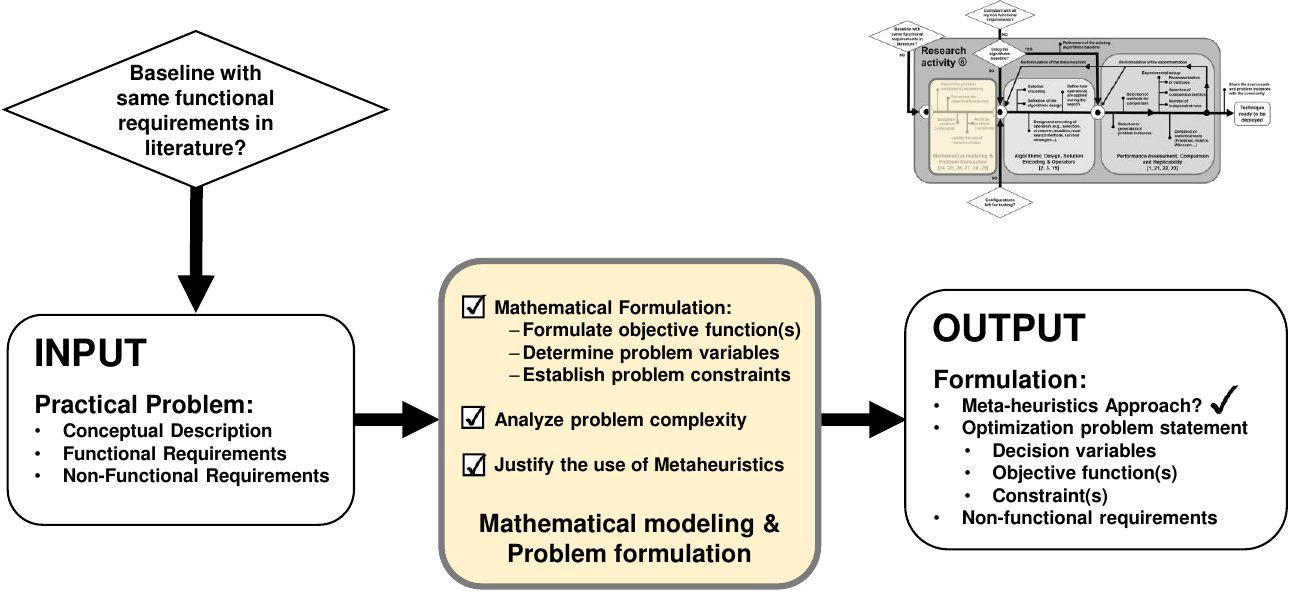}
\caption{Phase 1 of the reference workflow for solving optimization problems with metaheuristic algorithms.}
\label{fig:section4}
\end{figure} 
\subsection{Mathematical Formulation} \label{sec:mathForm}
The key steps to be followed are depicted in Figure~\ref{fig:section4} aiming at adequately translating a problem conception on paper into a precise mathematical formulation of an optimization problem:
\begin{itemize}[leftmargin=*]
    \item Clearly state the objective/cost function $f(\textbf{x})$ in charge of covering functional requirements and measure the quality and success of each assignment or solution. Try to infer as well the main characteristics of $f(\textbf{X})$: linear/nonlinear, unimodal/multimodal, or continuous/discontinuous. In multi-objective scenarios, when the decision functions are conflicting, multiple criteria will play a part in the decision making process. Nonetheless, the stakeholder might narrow down the Pareto optimal/nondominated solutions by imposing some preferences. 
    \item Characterize the decision variables ($\mathbf{x}=\{x_1,x_2,\ldots,x_n\}$) to be tuned (the order of the cities to visit, the community in which a node should be placed, the value of some input parameters,...) and their domain. 
    \item Determine the constraints of the problems as well as the natural or imposed restrictions, whose intersection will yield the feasible region of solutions. For Constrained Optimization Problems, hereafter called COPs, the nature of such constraints (equality, inequality or both) may be decisive in the algorithm approach selection.
\end{itemize}

Also, in Figure~\ref{fig:section4}, and for the sake of understandability, we have depicted in the upper right corner the placement of this phase in the Research Activity workflow (Figure~\ref{fig:workflow2}). The researcher will strive to accommodate the itemized list of functional and non-functional requirements into the mathematical formulation since setting boundaries to the automatic solution generation relieves subsequent efforts in modeling them at a next stage. We propose here a list of most common real-world non-functional requirements and some good practices to be assessed, if applied, by the researcher: 
\begin{itemize}[leftmargin=*]
\item \textit{Time consumption}. It is often the most relevant non-functional requirement expelled to take into account since the beginning of the mathematical conceptualization. 
    \begin{itemize}[leftmargin=*] 
        \item The fitness/objective function $f(\textbf{x})$ evaluation might be extremely time-consuming, specifically when equations are large and must be assessed in heavy computer-based simulations. Reformulations such as those approaches based on approximation-preserving reduction, i.e., relaxing the goal from finding the optimal solution to obtaining solutions within some bounded distance from the former \cite{approx1996}, surrogate objective functions \cite{lange2000optimization} or dimension reduction procedures (rightly after introduced) might be practical alternatives.
        \item Dimension reduction relates decision variables, the parameters on which the algorithm will perform the decision-making procedure. The length of such list $n=|\mathbf{x}|$ and their flexibility is strictly related to the time consumption required by the metaheuristic to explore the search space and run evaluations (i.e., $f(\textbf{x})$). Therefore, a preliminary study on the input parameters' selection, similar to Attribute Selection in Machine Learning, is strongly advocated in realistic scenarios oriented to real-world deployment. A parameterized complexity analysis might trigger a mathematical reformulation after delving into both the sensitivity of the objective function with respect to parameters \cite{Spagnol2019} (analogously to Information Gain in Machine Learning) and the inter-relation/correlation of each pair of input variables. The major concern about time consumption is likely to entice the researcher to pay close attention to the balance between problem dimensionality reduction and solution quality.
        \item Constraints may contribute to a faster convergence by narrowing the search in the feasible space. Nevertheless, the number of constraints (and their complexity) can also have a big impact on the existence of a solution and/or on the capacity of a numerical solver to find it. In fact, for real-life optimization problems, inequality constraints (physical limitations, operating modes, ...) can be quite large in comparison to decision variables $\mathbf{x}$, hence causing the feasible space to be shrunk to the point of eliminating any available solution. In such a case, the COP goal will be mathematically reformulated as finding the least infeasible vector of variable values.
    \end{itemize}
    
\item \textit{Accuracy of the solution}. Generally tightly related to the time-consumption requirement, once the mathematical formulation has been inferred, the optimization problem can be categorized into a convex (i.e. the objective function $f(\textbf{x})$ is a convex function and the feasible search space is a convex set) or non-convex one, which will mostly lead the algorithm selection process and its design. Researchers must get a balance between the aforementioned time consumption and the accuracy of the solution, especially on large scale non-convex spaces: are local optima acceptable results in favor of the computation lightening? are global optima achievable and verifiable in the real-world environment?. These questions will also flourish in the subsequent stages.

\item \textit{Unexpected algorithm interruptions must return feasible solutions}. In real-world environments, many unforeseen events may justify a need for a solution before the algorithm meets the stopping criteria thus finishing the search process. The solution, albeit premature, must be complete and fully compliant with the hard constraints. In such circumstances, the tendency to convert non-linear constraints into penalties (soft constraints) in the objective function to bias the solutions towards the frontiers is not a viable option.   
\end{itemize} 

With such an enumeration of requirements in hand, researchers should check those regarded at this initial stage and those not plausible for being satisfied by the mathematical formulation, which will be consequently transferred to the following adjacent phase.

        

\subsection{Analyze Problem Complexity - Justify the Use of Metaheuristics} 


Once the mathematical formulation has been completed, the research team involved in the work at hand should justify with solid grounds the need for metaheuristics to solve it efficiently. Mathematical optimization is a long-standing discipline, which has been traditionally focused on convex objective functions and feasible sets \cite{boyd2004convex}. Fortunately, there are already specific solvers (either exact or heuristics) suited to deal with this family of optimization problems, even to optimality when some specific conditions hold (e.g., linearity). Convexity ensures that every local optimum is a global optimum, hence avoiding common issues that motivate the use of heuristic and metaheuristic alternatives.

Unfortunately, the majority of contributions addressing real-world optimization problems just neglect any discussion on the convexity and properties of their mathematical formulations. Instead, they directly resort to the use of metaheuristics, without any major discussion on whether they are really needed \cite{sorensen2018history}. In this context, any prospective work along this line should pause at the following research questions:

\begin{itemize}[leftmargin=*]
\item \emph{Are the objective function(s) and constraint(s) analytically defined?} Intuitively, certain real-world optimization scenarios do not allow for an analytical formulation of the optimization problem itself. Indeed, the complexity of systems and assets to be optimized (as occurs in e.g., industrial machinery) jeopardizes the derivation of closed-form formulae for the objectives and constraints to be dealt with. However, this does not imply that quality and feasibility cannot be evaluated for any potential solution, but rather that the system/asset at hand must be considered as a black-box model that can be queried for any given input (solution). In this case, when the use of algorithms that do not depend or rely on the problem's properties becomes properly justified, it paves the way for the use of metaheuristic algorithms.

\item \emph{Can the problem be modified or reformulated without compromising the imposed functional requirements?} When the problem can be analytically defined, it might fail to comply with the mathematical properties that could allow exact methods and ad-hoc heuristics to be applied. For instance, even if the convexity of the objective(s) can be guaranteed, their linear or quadratic nature with respect to the optimization variables plays a crucial role in the adoption of exact linear and quadratic programming methods rather than optimization heuristics (e.g., gradient-based methods). At this point, it is strongly advised to examine strategies to reformulate (relax) the problem and mathematically simplify its objective(s) and constraint(s). These strategies include, among others, quadratic and linear transformations, constraint approximation via trust regions or Lagrangian relaxation \cite{ponton2017}. 

When considered and successfully applied to the problem at hand, the compliance of the reformulated problem concerning functional requirements should be analyzed. For instance, if the objective(s) are modified, a quantitative analysis of the implications of such modifications in the landscape of the original problem should be performed, particularly in regards to quality degradation (fitness value) and feasibility (constraint satisfaction). Depending on the chosen problem relaxation strategy, the reformulation could just penalize with an additive objective term those solutions as per their compliance with the imposed constraints. This reformulation is a crucial aspect that can be detected and it must be held in mind in subsequent design phases, as there is no mathematical guarantee that a feasible solution will be obtained. A similar conclusion can be drawn with linear relaxation strategies: is the quality (fitness) of the global optima of the relaxed problem far away from that of the original, unrelaxed problem? If there is an optimality gap, is it relevant for the real-world application under study? Unless these discussions are elaborated at this point of the reference workflow, design choices in subsequent phases can be made on the basis of a problem statement uncoupled from the requirements of the real-world problem under consideration.

\item \emph{Is the problem complex enough to discard simpler heuristics?} An equally relevant aspect of real-world problems is its complexity, which has been lately studied in the literature under the concept of \emph{fitness landscape} \cite{wright1932roles,reidys2002combinatorial,pitzer2012comprehensive}. In the context of optimization, fitness landscape comprises three essential elements of study: search space, fitness function, and neighborhood among solutions. Interestingly, since the search space and the definition of a neighborhood depend stringently on how solutions (optimization variables) are represented, the mathematical statement of the optimization problem and the algorithmic design of the solver to address it become entangled with each other \cite{merz1999fitness}. In other words, a single problem statement can span different fitness landscapes depending on how solutions are represented, even if dealing with continuous search spaces. The point is that only by assessing all these elements jointly, one can find solid reasons to opt for simpler heuristics, such as implicitly enumerative methods that rely extensively on the domains of study of landscape analysis (e.g., exhaustive search, Montecarlo sampling, neighborhood search, A* and branch and bound among others). Besides, landscape analysis can unveil other features with important implications that can be equally identified, such as ruggedness, basins of attraction, and funnels, to mention a few. When addressing real-world optimization scenarios with analytically defined problem formulations, we advocate for a closer look at these tools that, unfortunately, are often overseen in the literature related to real-world optimization. 

\item \emph{Is there expert knowledge about the problem/asset that should be considered in the definition of the problem?} In real-world settings, years of unassisted problem solving by users often accumulate expert knowledge that can be exploited in the design of efficient heuristics, as typically done by local search methods in memetic algorithms. However, we emphatically underscore the relevance of expert knowledge in terms of problem analysis. For instance, large regions of the search space can be discarded as per the experience of the user consuming the solution provided by the algorithm (implicit experience-based constraints). Likewise, the usability of the output in real application contexts can give valuable hints about how the problem can be relaxed, either in terms of formulation or in what refers to aspects impacting on its landscape (e.g., solution encoding, or how solutions can be compared to each other -- neighborhood). Section~\ref{sec:design} will later revolve on the capital role of expert knowledge in the design of optimization algorithms for real-world problems. However, this relevance also permeates to the definition of the problem itself and its eventual reformulations.
\end{itemize}

Summarizing the above points: metaheuristics must not be simply regarded as a \emph{swiss knife} for solving real-world problems, nor should this family of solvers be unduly applied to problems that can be simplified or tackled with simpler optimization methods. Instead, metaheuristics are powerful algorithmic enablers to deal efficiently with those cases of study whose complexity calls for their adoption. The provision of undeniable arguments for the necessity of metaheuristics should be enforced in prospective studies.

\section{Algorithmic Design, Solution Encoding and Search Operators} 
\label{sec:design}

\begin{figure}[h]
\centering
\includegraphics[width=1.0\columnwidth]{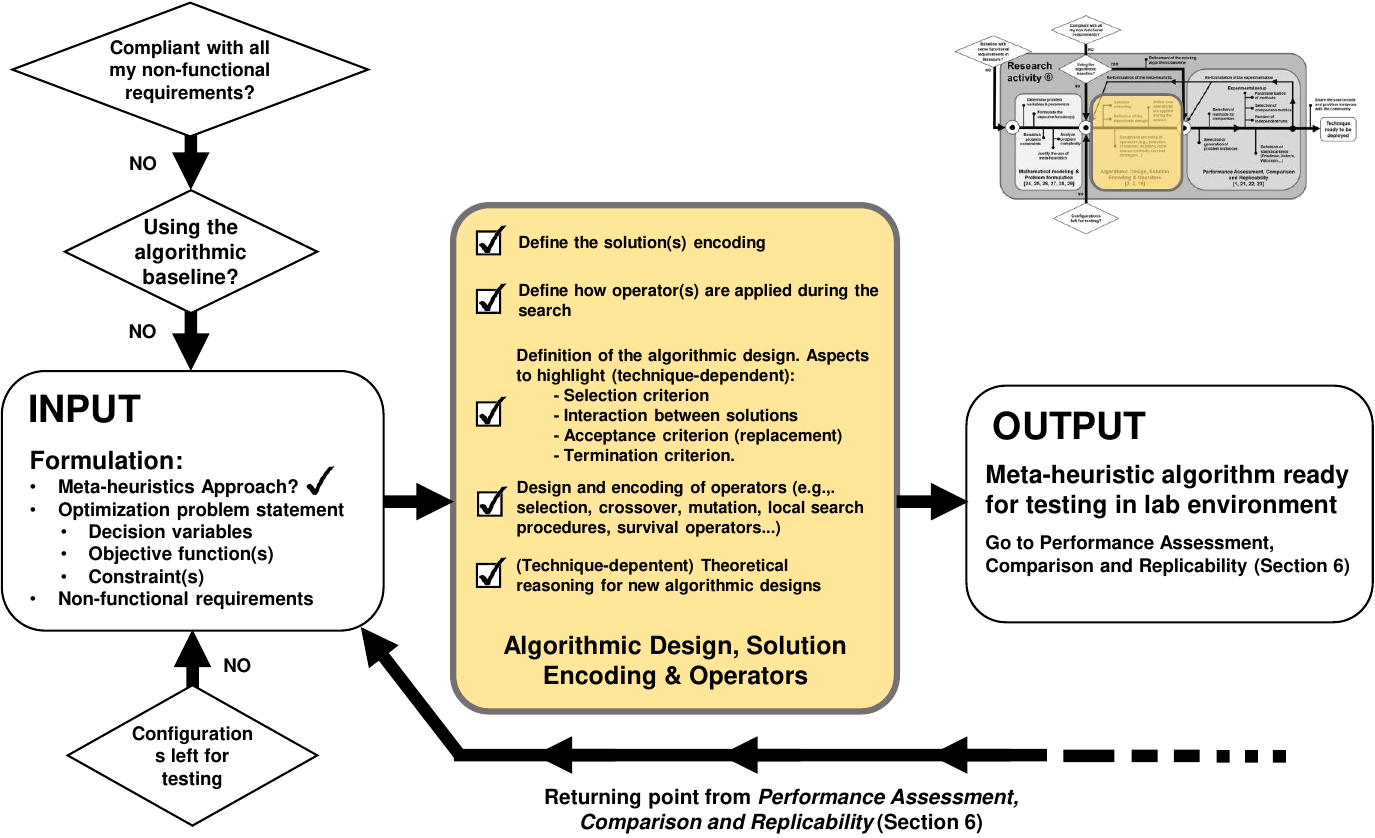}
\caption{Summary of the methodology on Algorithmic Design, Solution Encoding, and Search Operators.}
\label{fig:section5}
\end{figure}

The second phase to analyze after the problem modeling is the one devoted to the pure algorithmic design and development. Figure~\ref{fig:section5} summarizes the main aspects of this activity. As in the previous subsection, we have shown in the upper right corner the placement of this phase in the Research Activity workflow (Figure~\ref{fig:workflow2}). As can be seen in this scheme, and following the guidelines highlighted in the previous section, this step receives as input an optimization problem, formulated adequately as one or more objective functions, a set of decision variables, and a group of constraints. Furthermore, a group of must-fulfilling non-functional requirements is also provided as input, which undoubtedly influences the designs and developments conducted in this phase. This specific stage of the research can be reached from four different points:

\begin{enumerate}[leftmargin=*]
    \item Following the natural flow of research methodology depicted in Figure~\ref{fig:workflow1}, the metaheuristic design and implementation are conducted after the mathematical formulation of the problem (Section~\ref{sec:metho}).
    
    \item If researchers have found a baseline that meets the same functional requirements of the problem at hand, but the theoretical compliance of all non-functional requirements established (step \circled{5} in Figure~\ref{fig:workflow1}) cannot be verified.

    \item The re-design and re-implementation of the selected metaheuristic is necessary if the experiments carried out in the Lab Environment using a previously implemented solver do not verify the compliance of the defined functional requirements (Section~\ref{sec:assesment}).
    
    \item From the \textit{Algorithmic Deployment for Real-World Applications} step 
    (Section~\ref{sec:deployment} and \circled{7} in Figure~\ref{fig:workflow1}), only if the previously deployed solver does not meet the established non-functional requirements.
\end{enumerate}

Thus, these are the most important aspects a researcher or a practitioner should consider regarding the algorithmic design, solution encoding, and search operator development:

\begin{itemize}[leftmargin=*]
    \item \textbf{Solution encoding}. This is the first crucial decision to take in the algorithmic design \cite{ronald1997robust,talbi2009metaheuristics}. The type of encoding for representing the candidate solution(s) should be decided (real or discrete; binary \cite{chakraborty2003analysis}, permutation \cite{bierwirth1996permutation}, random keys \cite{bean1994genetic}, etc.). Its length (understood as the number of parameters that compose the solution) is also an essential choice. This length can be dependant on the size of the problem, or on the number of parameters to optimize. Thus, depending on these choices, encoded solutions can adopt different meanings. For example, the candidate can represent the problem solution itself (when genotype = phenotype \cite{rothlauf2006representations}), as in the case of the permutation encoding for the TSP \cite{larranaga1999genetic}, or a partial solution, as normally happens when using Ant Colony Optimization \cite{dorigo1999ant,blum2002ant}. Nonetheless, the candidate can represent a set of values acting as input for a specific system or a configuration of a defined set of preferences \cite{osaba2018multi} which will subsequently play a part in the complete problem solution. Taking this particularity into account, it is important not only to match the encoding to the problem (genotype vs phenotype) but to clearly detail it. For this reason, two important questions a researcher should answer are: ``Do we need to encode an individual for representing in a straightforward manner the problem's solution? Or we need an intermediate encoding better suited to test different heuristic operators?''.
    
    Focusing our attention on solutions encoded as parameters that act as inputs for an external system, researchers should bear in mind that the length of the candidate solutions and the domain of their variables are strictly related to the running times needed by the metaheuristic to modify and evaluate them. This impact on the running times is the reason for which, as mentioned in the previous section, a preliminary study on the input parameters to be considered is required for studies oriented to real-world deployment. This way, researchers could definitely choose which parameters should be part of the solution encoding, balancing both time consumption and influence in the solution quality. A remarkable number of studies have been published in the literature delving into this topic \cite{salcedo2002feature}, being the \textit{restricted search} mechanism \cite{salcedo2004enhancing} and the \textit{compression–expansion} \cite{salcedo2007improving} two representative strategies of this sort.
    
    Furthermore, the importance of solution encoding is twofold. On the one hand, it defines the solution space in which the solver works. On the other hand, the movement/variation operators to consider are dependant on this encoding. Consequently, different operators should be used depending on the encoding (e.g., real numbers, binary or discrete). Ideally, this representation should be wide and specific enough for representing all the feasible solutions to the problem. Additionally, it should fit at best as possible the domain search of the problem, avoiding the use of representations that unnecessarily enlarge this domain. In any case, and taking into account that this methodology is oriented to real-world deployments, non-functional requirements should be decisive for deciding which encoding is the most appropriate to deal with the problem at hand. For example, if the real environment contemplates unexpected algorithm interruptions (concept defined in Section~\ref{sec:mathForm}), encoding strategies allowing for partial solutions should be completely discarded. Moreover, if the execution time is a critical factor in the real system, representations that require a complex and time-consuming transformations or translations should also be avoided. An example of this translations is the Random-Keys based encoding, often used in Transfer Optimization environments \cite{gupta2015multifactorial,gupta2017insights}.
    

        

    
    \item \textbf{Population}. On the one hand, if the number of candidate solutions to optimize is just one, as in Simulated Annealing (SA, \cite{kirkpatrick1983optimization}) and Tabu Search (TS, \cite{glover1998tabu}), we can consider the metaheuristic as a trajectory-based method. On the other hand, if we deal with a group of solutions, the technique is classified as population-based. Examples of these solvers are GA and PSO. An additional consideration is the number of populations, which can also be more than one. These methods can be called multi-population, multi-meme, or island-based methods, depending on their nature \cite{alba2005parallel,alba2013parallel}. Instances of these approaches are the Imperialist Competitive Algorithm \cite{atashpaz2007imperialist} or the distributed and parallel GAs \cite{luque2011parallel}. In these specific cases, the way in which individuals are introduced in each sub-population should be clearly specified, and the way in which solutions migrate from one deme to another must also be formulated \cite{cantu1998survey}. Finally, well-known methods such as Artificial Bee Colony \cite{karaboga2007artificial} and Cuckoo Search (CS, \cite{yang2009cuckoo}) are characterized for being multi-agent, meaning that each individual of the community can behave differently.  
    
    Summarizing, the number of solutions to consider, the structure of the population, and the behavior of the individuals are three aspects that must be thoroughly studied. As in the previous case, non-functional requirements need to be carefully analyzed for making the right decision. For example, if the solver is run in a distributed environment, a multi-population method or a distributed master-slave approach (both synchronous or asynchronous) could be promising choices. Moreover, if the running time is a critical aspect and the problem is not expensive to evaluate, a single point search algorithm could be considered. In this regard, functional requirements must also be analyzed for choosing the proper alternative. For instance, if the solution space is non-convex and the number of local optima is high, a population-based metaheuristic should be selected, since it enhances the exploration of the search space. This aspect can be particularly observed in multimodal optimization \cite{das2011real,yang2009firefly}. 
    

        

    
    \item \textbf{Operators}. The design and implementation of operators is an important step that should also be carefully conducted. A priori, it is not a strict guideline for the development of functions in functional terms. Furthermore, there are different kinds of operators, such as selection, successor, or replacement functions, among others \cite{sivaraj2011review, blum2003metaheuristics,talbi2009metaheuristics}. In any case, and in order to avoid any ambiguity related to the terminology used \cite{prakasam2016metaheuristic,sorensen2015metaheuristics}, the way in which individuals evolve along the execution should be detailed using a standard mathematical language \cite{del2019bio}. In order to do that, each operator's inputs and outputs should be described using both algorithm descriptions and standard mathematical notation. We should also describe the nature of the operators (search based, constructive...) and the way in which they operate. Furthermore, and with the ambiguity avoidance in mind, it is advisable to anticipate possible resemblances with other algorithms from the literature and highlight differences (if any) by using, once again, mathematically defined concepts. 
    
    For example, a mutation operator of a GA can be mathematically formulated as:
    \begin{equation}
    \mathbf{x}^{t+1} = f_i(\mathbf{x}^t,Z)\in \mathcal{F},
    \end{equation} 
    where $\mathbf{x}^{t+1}$ is the output solution, and $Z$ denotes the number of times one of the functions $f_i()$ in $\mathcal{F}$ is applied to the input $\mathbf{x}^t$. Following the same notation, a crossover could be denoted as $\mathbf{z}^{t+1} = g_i(\{\mathbf{x}^t,\mathbf{y}^t\},Z)\in \mathcal{G}$.
    
    Again, non-functional requirements should be carefully studied to accurately choose or design all the operators that will be part of the whole algorithm. For example, some operators allow the eventual generation of incomplete and/or non-feasible solutions (i.e., solutions which do not meet all the constraints) to enhance the exploration capacity of the method. In any case, these alternatives should be avoided in case the real-world scenario considers unexpected algorithm interruptions. Additionally, in case the running time is a critical issue, operators that favor the convergence of the algorithm should be prioritized (understanding convergence as the computational effort that the algorithm requires for reaching a final solution(s) \cite{olafsson2006metaheuristics}). 


        

    
    \item \textbf{Algorithmic Design}. Briefly explained, the algorithmic design dictates how operators are applied to the solution or groups of solutions. It could be said that this design determines the type of metaheuristic developed. At this point, it should be mandatory to provide overall details of the algorithm. To do this, several alternatives are useful, such as a flow diagram, a mathematical description or a pseudocode of the method. Furthermore, if the modeled technique incorporates any novel ingredient, it is highly desirable to conduct this overall description of the method using references to other algorithmic schemes made for similar purposes. Furthermore, the number of possible alternatives for building a solution metaheuristic is really immense, being impossible to point here all the aspects that should be highlighted. In any case, some of the facets that must be described are the selection criterion, the criterion for the interaction among solutions (in terms of recombination in GAs, or migration in multi-population metaheuristics), the acceptance criterion (replacement) and the termination criterion.
    
    Probably, the first good practice to follow when deciding the algorithmic design of a real-world oriented metaheuristic is to take a detailed look at recent related scientific competitions. Tournaments such as the ones celebrated in reference conferences such as the {\em IEEE Congress on Evolutionary Computation} \footnote{\url{https://www.ntu.edu.sg/home/epnsugan/index_files/CEC2020/CEC2020-1.htm}} and the {\em Genetic and Evolutionary Computation Conference} should guide the selection of the candidate algorithm. For making this decision, it should be checked if the real-world problem belongs to a class of problems with a similar competition benchmark, being meaningful in this case to focus the attention on those algorithms that have shown a remarkable performance at recent competitions.
    
    Once again, researchers should thoroughly consider both functional or non-functional requirements for properly choosing the design of the metaheuristic. For example, computationally demanding designs could be acceptable only in situations in which the running time is not critical. On the contrary, if we want to reduce the execution time by sacrificing some quality in the solution, the termination criterion would be a cornerstone for reaching a proper and desirable convergence. Interaction between candidate solutions would also be of paramount importance if the implemented algorithms will be deployed in a distributed environment, requiring advanced and carefully designed communication mechanisms. 
    
    Regarding the problem complexity, if this is remarkably high, automated algorithm selection mechanisms can be an appropriate alternative \cite{kerschke2019automated}. This concept sinks its roots in the well-known no-free-lunch theorem \cite{wolpert1995no}. This theorem particularly applies in computationally demanding problems, in which no single algorithm defines the baseline. On the contrary, there is a group of alternatives with complementary strengths. In this context, automated algorithm selection mechanisms can, within a predefined group of algorithms, decide which one can be expected to perform best on each instance of the problem.
    
    Another interesting aspect to consider for the algorithmic design is the whole complexity of the technique. Usually, the development of complex algorithmic schemes is unnecessary, if not detrimental. Some influential authors have proposed the bottom-up building of metaheuristics, using the natural trial and error procedure. It has been demonstrated how, in practice, robust optimizers can be built, which can compete with complex and computationally expensive methods. This concept, based on the philosophical concept of Occam’s Razor, is the focus of some interesting studies such as \cite{iacca2012ockham,caraffini2012three}.
    
    An additional consideration that should be taken into account for properly choosing the algorithmic design is the expertise of the final user. In this sense, if the user who will use the deployed method in the real environment has no experience with these kinds of techniques, it is recommended to implement techniques needing a slight parameterization. Examples of these methods are the basic versions of the Cuckoo Search or Differential Evolution. Other promising alternatives for these types of situations are the solvers known as adaptive \cite{cotta2008adaptive}, or the automated design methods \cite{woodward2011automatically,woodward2012automatic}. On the contrary, if the final user is familiar with the topic, the researcher could deploy a flexible solver configurable by several control parameters to allow refinements in the future. Well-known examples of these methods are the Genetic Algorithm (with its crossover and mutation probabilities, population size, replacement strategy, and many other parameters) or the Bat Algorithm (with its loudness, pulse rate, frequency or wavelength, among other parameters).
    
    Furthermore, and although the interest in providing theoretical guarantees of newly developed metaheuristics is in crescendo, we should also explicitly call in this methodological paper for an effort to incorporate theoretical reasons for new algorithmic designs. In other words, we should progressively shift from a performance-driven rationale (\textit{look, my algorithm works}) to a theory-/intuition-driven design rationale (\textit{look, my algorithm will work because,...}). Of course, this trend should also be extended not only to the algorithmic design, but also to the generation of new operators and operation mechanisms. 
    
    Finally, and referring to the proposal of new metaheuristics, operators, or mechanisms, we want to highlight the importance of properly describing all the aspects involved in a solver using a standard language. In other words, all metaphoric features should be left apart, or contextualized using openly accepted methods as references. In fact, the lack of depth in these descriptions is the main reason for lots of ambiguities generated in the literature \cite{del2019bio,sorensen2018history}. For example, it is perfectly valid to name the individuals of a population as Raindrops, Colonies, Bees or Particles, but they must be notated using a standard mathematical language, and it should be clarified that they are similar to an individual of a Genetic Algorithm (if we use the GA as a reference).
    

        
        

    
\end{itemize}

\section{Performance Assessment, Comparison and Replicability} 
\label{sec:assesment}

When the selection of the algorithms is carried out by considering previous reports and studies in the literature, this step is indeed not needed. However, it is quite frequent that good comparisons do not exist in the literature to make a reasonable decision. This implies that we have to conduct our own comparisons in order to select the algorithm that better meets our requirements. This section discusses on several aspects that must be considered to conduct a rigorous and fair experimentation to make that decision. Specifically, these topics are: experimental benchmark \ref{sec:ben}, evaluation score \ref{sec:eva}, fair comparisons among techniques \ref{sec:rich}, statistical testing \ref{sec:stat}, and replicability \ref{sec:rep}. 

\subsection{Experimental benchmark} \label{sec:ben}

This is usually the first decision researchers must face when designing the experimental evaluation of their proposal. In this regard, depending on the kind of contribution (more theoretical or more applied) the nature of the problems used in the experimentation might be different. In the first case, researchers would probably use synthetic benchmark functions to assess the performance or the advantages of the different proposals or considered algorithms. In the second case, the authors would normally propose a set of instances related to the problem they are trying to solve. Sometimes, especially when they are dealing with a novel or a very specific problem with strong requirements, one has to design his own benchmark. In all the cases, those problem instances must comply with several conditions to ensure that the experiments succeed in assessing the significance of the decision of selecting one or another technique. In order to do that, benchmarks should be designed not only to allow the evaluation of functional but also of non-functional requirements, in order to analyze the degree to which all of them are met. Additionally, it should also be considered specific conditions allowing or encouraging the eventual application of statistical tests, such as a large number of problems and/or an odd number of them to reduce potential problems in the comparative analysis, due to the \textit{cycle ranking} or the \textit{survival of the nonfittest} paradoxes \cite{liuParadoxesNumericalComparison2020}. Since the methodology introduced in this paper is oriented to the deployment of real-world metaheuristics, we recommend conducting laboratory tests using datasets as real as possible, even if they are synthetically generated. 

In this regard, it is widely agreed in the community that real-world benchmarks have been traditionally scarce. However, significant efforts have been recently conducted to overcome this lack. In \cite{tanabe2020easy} an easy-to-use multi-objective optimization suite is introduced, consisting of 16 bound-constrained real-world problems. Similar studies have been also contributed in \cite{cheng2017benchmark} and \cite{chen2020proposal}, focused on realistic many-objective optimization problems. In \cite{picard2020realistic}, a generic framework is proposed to design geared electro-mechanical actuators. The proposed framework is utilized for constructing realistic multi-objective optimization test suite, with an emphasis placed on constraint handling. Another work along this line can be found in \cite{kumar2020test}, which describes a benchmark suite composed by 57 real-world constrained optimization problems. An additional proposal published in \cite{he2019repository} revolves on data-driven evolutionary multi-objective optimization. Additional studies of this kind can be found in \cite{lou2019constructing}, \cite{ishibuchi2019scalable}, and \cite{omidvar2015designing}. In any case, for the generation of valuable synthetic problem instances, all the variables that compose the real situation must be thoroughly studied in order to build reliable test cases. Thus, newly built datasets should be adapted to these variables. Lastly, if any real instance is available, the generation of additional synthetic test cases is recommended, using the real one as inspiration.

Furthermore, when the experimental benchmark is made up of synthetic functions, such functions should be a challenge for optimization algorithms. Thus, the benchmark should comprise functions of different nature and challenging features, such as a different number of local optima, shifting of the global optimum, rotation of the coordinate system, non-separability of (sub-)components, noise, several problem sizes, etc., depending on the expected features of the problem to tackle. Designing such a benchmark can be a difficult task, so a good recommendation is to use some of the existing benchmarks in the literature. The use of well-known benchmarks also facilitates the selection of the reference algorithms to be included in the comparison. It should be finally pointed out that, although a technique will be deployed in a real environment solving a real-world problem, conducting tests with this kind of datasets is of great importance for measuring the quality of the developed proposal.

On the other hand, when the experimental benchmark includes real-world problems that have not been tackled before, authors should carefully select an appropriate set of instances to evaluate their proposal, or even generate them. This last case is especially frequent in situations where the problem is being solved for the first time by the community, something frequent when dealing with real environments. Thus, the testbed should include a broad set of instances covering all the relevant characteristics of the problem under consideration in order to resemble, as much as possible, the real-world scenarios being modeled. These dataset should be chosen or generated for efficiently testing each functional and non-functional requirements of the problem. Lastly, instances should be described in detail and, whenever possible, should be made available to the community, so that other authors can use them to evaluate their own contributions.

\subsection{Evaluation measure} \label{sec:eva}

An optimization algorithm can be assessed from different points of view and based on many features. Traditionally, main measures are related to the performance (a \textit{fitness} function or an error measure). However, there are many other possible measures of interest in a real-world context:
\begin{itemize}[leftmargin=*]
\item \textit{Processing time}, which depends on computational complexity. In many real-world contexts this response time is crucial. In this context, it is recommended to generate a record containing the execution times presented by all the considered metaheuristics. This record should be associated with the computational environment in which techniques have been run. Thus, this logbook will be useful in the subsequent deployment phase, and it is specially crucial for properly measuring the impact of the system migration on the algorithm's running time. 

\item \textit{Memory requirements}: This is especially important when the algorithm is expected to run in hardware with limited resources.

\item For distributed algorithms there are special measures such as the \textit{communication latency}, or the \textit{achieved speedup} (relative to the number of nodes). Additionally, other non-functional requirements can also be considered and measured, such as robustness when a node fails, redundancy, etc.

\item The \textit{required time to obtain a reasonably good solution}, especially in problems in which each evaluation requires significant computational resources. In these scenarios, algorithms often apply surrogate models to reduce their execution time.
\end{itemize}

As a general rule, the assessment of the performance of an optimization algorithm can not be guaranteed if the measure of just a single run is reported. Robust estimators of an evaluation metric can only be computed if enough information is available. In this sense, multiple runs should be considered so that the statistical methods described below can deliver significant conclusions. Special attention should also be paid to the fact that multiple runs must be independent, i.e., no information is fed from one run to another.

\subsection{Rich comparisons from multiple perspectives} \label{sec:rich}
A rigorous assessment of an optimization method should focus on different aspects of the method behavior, and it should explore different perspectives for gathering meaningful insights. For example, aligned tables with min, max and mean results should only be considered for informative purposes. Nevertheless, much richer visualizations should be analyzed to dive in the data and highlight the most important findings of the research. One possible approach to visualize the comparisons between algorithms is the use of data profile techniques like the one proposed in \cite{moreBenchmarkingDerivativeFreeOptimization2009}, which was later extended in \cite{liuBenchmarkingStochasticAlgorithms2017}. The modified data profile technique proposed in these studies allow comparing several optimization algorithms by adopting a two-step methodology: a comparison of the mean in the first step, and a comparison of confidence bounds in the second phase.

In some real-world problems, specific visualizations can be helpful to ease the interpretation of the results of the optimization algorithm. For example, in a routing problem, the visualization of the routes can be examined by an expert in mobility that will be able to assess the convenience of using the solutions provided by the optimization algorithm. In some cases, a route with a longer distance might be more appropriate if it complies with some additional constraints that were not available when the problem was defined than a route with a shorter distance which violates such constraints. This can be easily spotted by the expert with this kind of visualizations.

Something similar is recommended for real-parameter optimization problems. In this case, it is also useful to depict different solutions, but an alternative visualization should be considered. This visualization should make possible to represent not only the solutions themselves, but also the fitness value associated to each solution, in order to identify promising regions of the solution space. A direct approach for visualizing continuous variables would be to use 2D or 3D scatter plots, in the case of very small problems. If the problem has 4 variables or more, it is not possible to represent solutions without the use of dimensionality reduction techniques (PCA, t-SNE, UMAP, etc.). An alternative approach is the use of parallel coordinate plots. 

In the case of multi-objective optimization problems, they also require specific visualization techniques. In this context, it is of great interest to be able to represent the Pareto front of the optimization problem to allow the user to choose from among all the available non-dominated solutions. 

%
%

Another important issue that should be subject of study is how algorithms manage the exploration vs. exploitation ratio \cite{LaTorre2010d,Herrera-Poyatos2017}. In most cases, authors do not put significant attention on how the components of the developed techniques contribute to exploration/exploitation. However, no analysis to support this hypothesis is normally carried out, and such analysis should be mandatory \cite{Crepinsek2013}. 


Another crucial aspect, which has been also mentioned in previous sections, deals with the complexity of the algorithms. In this sense, an intuitive approach is to compare the running times of the algorithms under study. However, this measure is only meaningful in certain real-world situations. Other elements could also affect this performance measure: differences in the computing platform, availability of a parallel implementation, the application of the code, etc. For this reason, other language-agnostic measures such as the Cyclomatic Complexity (or Conditional Complexity, or McCabe's Complexity) \cite{McCabe1976}, are normally preferred. More concretely, Cyclomatic Complexity is a software metric that measures the number of independent paths in a program source code. The higher the number of independent paths are, the more complex the program is and, thus, a higher complexity value is obtained. Nonetheless, the efficiency of the algorithm, in terms of their consumption of computing resources, can be of utmost importance for real-world oriented research. 

The last fundamental feature pointed out in this subsection relates to the adjustment of the parameter values of each algorithm. In this sense, it makes sense to adjust the parameter values to adapt the search to the complexity of the instance/problem, given that this complexity can be directly inferred from the information that we have of the instance/problem (such as, for example, its size), without the need of additional processing to identify it. If a parameter tuning algorithm has been employed (which is highly recommended, see \cite{del2019bio}), the tuned values should also be analyzed. An additional aspect to consider is to clearly analyze the influence of each parameter in the fulfillment of established functional and non-functional requirements, and to analyze the impact of the fine-grained tuning of each parameter value. The depth comprehension of this influence is of great value for providing a sort of understandability framework to non-familiarized stakeholders. In this regard, algorithm developers should prioritize techniques and systems that can be parameterized externally, so that such parameterization can be carried out by non-experts in the field.

\subsection{Statistical Testing} \label{sec:stat}
\label{sec:statistics}

Statistical comparisons of results should be considered mandatory, especially when the algorithms used in the experimentation are stochastic. However, even if the statistical comparisons are made, they are not always correctly carried out. There are some popular methods in inferential hypothesis testing, such as the t-test or the ANOVA family. Nonetheless, these tests, called \emph{parametric tests}, assume a series of hypotheses on the data to which they are applied (normality, homocedasticity, etc.). If those assumptions do not hold, their reliability is not guaranteed, and alternative approaches should be considered. This is the case of \emph{non-parametric} tests, such as Wilcoxon's test, which do not assume any particular characteristic of the distribution of the underlying data \cite{demsarStatisticalComparisonsClassifiers2006}. Consequently, these tests can be more generally applied. However, they are less powerful than parametric tests as they consider the relative ranking instead of the real error values of the different proposals.

Additionally, when several comparisons are done, the cumulative error should be carefully considered. For instance, the popular Wilcoxon's test is a test designed for comparing two data samples (usually coming from the errors of the algorithms subject to comparison). When more than two samples are compared among them, the cumulative error could increase \cite{derrac2011practical}. In these cases, a post-hoc treatment such as Holm (or others) should be used to keep this cumulative error under control in the overall comparison.

It should be noted, however, that the use of statistical tests does not guarantee that errors in the interpretation of results will not occur. Indeed, the concept of p-value can lead to several misinterpretations. This same problem could also arise when using confidence intervals methods, but it has been proven that it happens in a smaller scale \cite{greenlandStatisticalTestsValues2016}. Also, in \cite{liuParadoxesNumericalComparison2020} two popular comparison strategies are analyzed, obtaining several paradoxes that could lead to different misinterpretations of results. In particular, comparing by pairs of algorithms, as it is done when using well-known t-test or Wilcoxon's test, could produce the \textit{cycle ranking} paradox, concluding that none of the compared algorithms could be identified as the best one. Furthermore, methods like ANOVA, which compares multiple algorithms, may lead to the \textit{survival of the non-fittest paradox}, by which the identified winner could differ from the one obtained through statistical comparisons of pairs of algorithms.

The above inferential tests are based on frequentist statistics, and present several problems, the most obvious being the degree of dependence between the p-value and the confidence intervals with respect to the size of the sample. Generally, when enough data is available, it is very simple to obtain a small p-value. Since the sample size is arbitrarily chosen by the researcher and the null hypothesis (samples come from the same distribution) is usually wrong, the researcher can reject it by testing the algorithms with a larger amount of data. On the contrary, considerable differences could not yield small p-values if there are not enough data (as datasets) for testing the method \cite{bayes}. More recently, the use of Bayesian statistical tests is attracting more and more interest, as they are considered to be more stable and the interpretation of their results is more appropriate to what researchers want to analyze \cite{bayes}.

\subsection{Replicability of the Experiments} \label{sec:rep}
\label{sec:replicability}

As the last point of this section, replicability is one of the standard criteria used to assess the scientific value of a research. With replication, different and independent researchers can address a scientific hypothesis and build up evidence for or against it. 
In this section, we are going to describe different considerations that should be taken into account to make possible the replicability of experiments.

A good practice for comparing multiple algorithms is to ensure that all of them have been configured by following the same approach, with the same exhaustiveness, so all of them are run on the same environment and experimental conditions that guarantees that none of them has an advantage over the others. This is a crucial aspect for determining which approach will perform better in the real environment. In order to accomplish that, it is important to use a benchmark that does not have an unfair bias which favors some algorithms over the others. This is particularly important when the algorithms are tested using a synthetic benchmark, because it could have some characteristics that are uncommon in real-world problems. 
As we are concerned about real-world problems, any specific feature that could be useful in any particular optimization algorithm would be of particular interest.

Regarding the experimental conditions, another important issue that should be taken into account is the maximum processing time, which is strictly determined by the real-world problem to be solved. A good practice in this context is the allocation of a dedicated budget of objective function evaluations for each of the algorithms in the experimentation. This budget should be determined by an estimation of the time complexity required by each algorithm in the benchmark. In turn, the estimated time complexity of a method should be subject to the implementation that would be subsequently deployed over the real environment. Also, it is advisable to perform quantitative time complexity assessments for each of the stages that comprise the whole metaheuristic technique. It should also be clear that the time complexity can be influenced by multiple factors, including the hardware in which the experiments are run and the software of the implementation in use, such as the operating system, the programming language and/or the compiler/interpreter. A change in any of these factors could significantly alter the performance estimation of the algorithms under comparison. In any case, the maximum processing time of an algorithm is an important decision driver that has to be considered when designing the algorithmic solution. Otherwise, the selection of one metaheuristic approach over other possibilities could be of no practical use when applied to the real-world scenario under study.

All the previous requirements are needed to guarantee the replicability of the experimental conditions. However, we can not talk about replicability if the specific instances/problems used in the experiments are not readily available to external researchers. In the specific case of this methodology, which is oriented to real-world applications, it could be possible that the instance/problem used contains internal and private data, which should not be shared publicly. In these situations, researchers should comply with the corresponding legal limitations before the public sharing of the data (such as anonymizing private data, for example). Additionally, it could be interesting to provide connectors for different languages and/or frameworks. Furthermore, if the problem datasets have been generated synthetically (as mentioned in 
Section~\ref{sec:ben}), it is highly recommended to publicly share the instance generator that was adopted.

Finally, and although it is usually not considered a requirement, making the source code of a new algorithm freely available to facilitate replicating the results is highly recommended. In this regard, and depending on the context, confidentiality issues can arise between the algorithm developer and the stakeholder. In this case, several actions can be conducted, such as the anonymization of the code, or the generalization of the method. Thus, the principal reason for enhancing the sharing of the source code is that, very often \cite{biedrzyckiEquivalenceAlgorithmImplementations2019}, many details in the implementation that have a strong influence on the results are not included in the descriptions provided. Thus, without a reference implementation, many implementations of the same algorithms could deeply differ in their results. Thus, the source code should be shared in a permanent and public repository, such as GitHub, Gitlab, Bitbucket, etc., to name a few. If confidentiality is a problem, a contact e-mail could be shared for code sharing requests.

The conjunction of the availability of both the data and the source code of the algorithm is what is called \textit{``Open Science''} \cite{killeenPredictControlReplicate2019,pengReproducibleResearchComputational2011,collaborationReproducibilityProjectModel2013}, and it is an increasingly popular approach to ensuring replicability in science, so that we can make better and better science.

\section{Algorithmic Deployment for Real-World Applications}
\label{sec:deployment}



Once we have completed the steps of the Lab Environment phase of Workflow \ref{fig:workflow1}, it's time to proceed with the second part of Figure~\ref{fig:workflow1}, namely the Application Environment, which is focused on the algorithmic deployment for the real application at hand. As pointed out in  Section~\ref{sec:workflow}, this phase receives as input either an algorithm implementation taken from an existing software package or an ad-hoc method defined in the algorithmic design step of Workflow \ref{fig:workflow2}. In both cases, this implementation should go through a verification process to determine whether it fulfills the functional and non-functional requirements to be deployed in a real environment. If this is not the case, then a new implementation should be addressed.

When facing the development of a metaheuristic to be deployed in a real-world environment, several factors can lead to taking one of the following approaches: to implement the algorithm from scratch or to choose an existing optimization framework. Among these factors, we can consider:

\begin{itemize}[leftmargin=*]
    \item \textit{Programming skills}. If the development team has a high expertise then a choice is to determine whether it can afford to make an implementation from scratch. Consequently, the written code can be highly optimized, and thus it is more likely to meet the non-functional requirements, particularly those related to performance. The counterpart of this approach is that the code may be difficult to be updated, extended, and reused by other people (including the development team itself).
    
    \item \textit{Using an existing optimization framework}. The most productive approach to develop a metaheuristic is to build on existing frameworks. This way, most of the needed algorithmic components may be already provided, so there is no need to reinvent the wheel, and they can offer additional functionality (e.g., visualization, analysis tools, etc.). If a goal of the metaheuristic to be developed is to offer it to the community (in principle, as an open-source contribution), integrating it into a framework is probably the best choice. However, as a possible negative point, using a framework imposes the use of a set of existing base components, so the resulting implementation could not be as efficient as one developed ad-hoc, and thus non-functional requirements related to performance could be affected.
    
    \item \textit{Corporate development platform}. Many companies have a preferred software platform to develop their products, e.g., Java, .NET (C\#, Visual Basic), etc., which can impose constraints affecting the implementation of the algorithm, both in the sense of the optimization frameworks that could be used and the availability of third-party libraries. In this sense, programming languages such as Python are becoming very popular due to the large number of existing libraries for data analysis, visualization, and parallel execution.
    
    \item \textit{Software license}. An important issue to consider when using third-party software is the licensing policy. Some licenses, such as GPL (GNU General Public License ) or LGPL (GNU Lesser General Public License), can be too restrictive and thus can hinder the adoption of software packages in non-open-source applications. Others, including MIT and Apache, are less restrictive.
    
    \item \textit{Project activity}. If an existing software package is attractive to use, it is important to determine whether the project is still active, which ensures, at least in theory, the possibility of contacting the authors to report bugs found or to answer questions than are not included in the project's documentation. There is also the choice of requesting support for the project developers. 
    
\end{itemize}

\begin{table}[]
    \centering
    \resizebox{\columnwidth}{!}{
    \begin{tabular}{lccccc}
    \toprule
        {\bf Framework} & {\bf Language} & {\bf Algorithms} &  {\bf License} & 
        {\bf Current version} & 
        {\bf Last update} \\
        \midrule
        \multicolumn{1}{l}{ECJ \cite{ECJ19}} & Java & {\bf SO}/MO & AFL & 27 & August 2019 \\
        \multicolumn{1}{l}{HeuristicLab \cite{wagner2014}} & C\# & {\bf SO}/MO & GPLv3 & 3.3 & July 2019 \\
        \multicolumn{1}{l}{jMetal \cite{DN11}\cite{NDV15}} & Java & SO/{\bf MO} & MIT & 5.10 & July 2020 \\
        \multicolumn{1}{l}{jMetalCpp \cite{jmetalcpp13}} & C++ & SO/{\bf MO} & LGPL & 1.8 & November 2019 \\
      \multicolumn{1}{l}{jMetalPy \cite{jMetalPy19}} & Python & SO/{\bf MO}  & MIT & 1.5.3 & February 2020 \\
        \multicolumn{1}{l}{MOEAFramework \cite{Hadka20}} & Java & SO/{\bf MO} & LGPL & 2.13 & December 2019 \\
        \multicolumn{1}{l}{NiaPy \cite{NiaPyJOSS2018}} & Python & SO & MIT &  2.0.0 & November 2019 \\
        \multicolumn{1}{l}{Pagmo \cite{Pagmo20}} & C++ & {\bf SO}/MO & GPL/LGPL & 2.16 & September 2020 \\
        \multicolumn{1}{l}{ParadisEO \cite{Paradiseo04}} & C++ & {\bf SO}/MO &  CeCill & 2.0.1 & December 2018 \\
        \multicolumn{1}{l}{PlatEMO \cite{PlatEMO17}} & MATLAB & MO & Open source & 2.9 & October 2020 \\
        \multicolumn{1}{l}{Pygmo \cite{Pygmo20}} & Python & {\bf SO}/MO &  Mozilla & 2.16 & September 2020 \\
        \multicolumn{1}{l}{Platypus \cite{Platypus20}} & Python & SO/{\bf MO} & GPLv3 & 1.0.4 & April 2020 \\
        \bottomrule
    \end{tabular}}
    \caption{Main features of representative multi-objective optimization frameworks. ``SO/MO'' in column Algorithms stand for single-objective/multi-objective algorithms. If a framework provides both types of algorithms but it is more focused on one them, it is highlighted in {\bf boldface}.}
    \label{tab:optimizationframeworks}
\end{table}

Table~\ref{tab:optimizationframeworks} contains a summary of the main features of a representative set of metaheuristic optimization frameworks. The characteristics reported include the programming language used in the project, the main focus of the framework (most of them include single- and multi-objective algorithms, but they usually are centered on one of them), the software licence, and the current version and last update date (at the time of writing this paper). 

Attending to the programming language, we observe that Java, Python, and C++ are popular choices, but we also find HeuristicLab and PlatEMO, which are developed in C\# and MATLAB, respectively. At first glance, it might be assumed a priori that Python-based frameworks would be computationally inefficient, so if this is a non-functional requirement, then others based on C++ or even Java could be more appropriate. However, Pygmo is in fact based on Pagmo (it is basically a Python wrapper of that package, which becomes a drawback to Python users if the intend to use Pygmo to develop new algorithms), so it can be very competitive in terms of performance. The other frameworks written in Python are considerable slower; for example, if we consider jMetal (Java) and jMetalPy (Python), it can be seen that running the same algorithm with identical settings (e.g., the default NGSA-II algorithm provided in both packages) can take up to fifteen times more computing time in Python than in Java. In return, the benefits of Python for fast prototyping and the large number of libraries available for data analysis and visualization make the frameworks written in this language ideal for testing and fine-tuning. 

The orientation of the frameworks on single- or multi-objective optimization can be a stronger reason to choose a particular package than the programming language. Thus, if the problem at hand is single-objective, then ECJ, HeuristicLab, Pagmo/Pygmo, ParadisEO, or NiaPy offers a wide range of features and algorithms to deal with it. The same applies with the other frameworks concerning multi-objective optimization; in this regard, it is worth mentioning jMetal, which started in 2006 and it is still an ongoing project which is continuously evolving, and PlatEMO, which appeared a few years ago and offers more than 100 multi-objective algorithms and more than 200 benchmark problems.

The type of software licenses can be a key feature that may disable the choice of a particular package. For example, PlatEMO is free to be used in research works according to its authors, so it is not clear whether it can be used in industrial or commercial applications. In this regard, the first release of jMetal had a GPL license, which was changed a few years later to LGPL and, more recently, to MIT upon request of researchers working in companies that wanted to use the framework in their projects. 


When the metaheuristic has been implemented, it is advisable to perform a fine-tuning to improve its performance as much as possible. This process has two dimensions. First, the code can be optimized by applying profiling tools to determine how the computational resources available are distributed among the functions to be optimized. This way, code parts consuming considerable time fractions can be detected, and they can be refactored by rewriting them to make them more efficient. We have to note that metaheuristics consist of a loop where several steps (e.g., selection, variation, evaluation, and replacement in the case of evolutionary algorithms) are repeated thousands or millions of times, so any small improvement in a part of the code can have a high impact in the total computing time. 

The second dimension is to adjust the parameters settings of the algorithm to improve its efficacy, which can be carried out by following two main approaches: ad-hoc pilot tests and automatic configuration. The first approach is the most widely used in practice, and it is advisable when having a high degree of expertise; otherwise, it usually turns into a loop of trial and error steps lacking rigor and leading to a waste of much time. The second alternative implies the use of tools for automatic parameter tuning of metaheuristics~\cite{HLUY20}, such as irace~\cite{LDP+16} and ParamILS~\cite{HHL+09}, although it must be taken into account that the tuning with these kinds of tools can be computationally unaffordable in real-world problems. 


At this point, the new implementation should again be verified against the non-functional requirements, which could imply to review the implementation in case of not fulfilling some of them. If this is not the case, the metaheuristic may still not be ready to be used in a real environment because of the potential appearance of new non-functional requirements. This situation can happen due to a number of facts, such as the following:
\begin{itemize}[leftmargin=*]
    \item Changes in the deployment environment. The real system was not specified in detail when the problem was defined (e.g., the target computing system is not as powerful as previously expected), so there can be a requirement fulfillment degradation that was not observed in the in-lab development. 
    \item The client is satisfied with the results obtained by the metaheuristic, so it is applied to more complex scenarios than expected. Consequently, the quality of the solutions cannot be satisfactory, or time constraints can be violated.
    \item Once the algorithm is running, the domain expert notices new situations that were not taken into account when the functional and non-functional requirements were defined.
    \item The algorithm is not robust enough, and there may be significant differences in the obtained solutions under similar conditions, which can be confusing for the user.
    \item In the case of multi-objective problems, providing an accurate Pareto front approximation, with a high number of solutions, can overwhelm the decision maker if it is merely presented. The algorithm could be empowered then with a high-level visualization component to assist in choosing a particular solution (a posteriori decision making). Even a dynamic preference articulation mechanism could be incorporated to guide the search during the optimization process (interactive decision making).
\end{itemize}

If the metaheuristic is not compliant with all the new non-functional requirements, it must be analyzed whether they can be fulfilled by re-adjusting the parameters settings or by carrying out a new implementation; on the contrary, it can be necessary to go back again to the research activity or even to the problem description.


        



\section{Summary of Lessons Learned and Recommendations} \label{sec:recommendations}

The final purpose of the methodology discussed heretofore is to avoid several problems, poor practices and practical issues often encountered in projects dealing with real-world optimization problems. As a prescriptive summary of the phases in which the methodology is divided, we herein provide a set of synthesized recommendations that should help even further when following them in prospective studies. Such recommendations are conceptually sketched in Figure~\ref{fig:recommendations}, and are listed next:
\begin{figure}[h!]
\centering
\includegraphics[width=1.0\columnwidth]{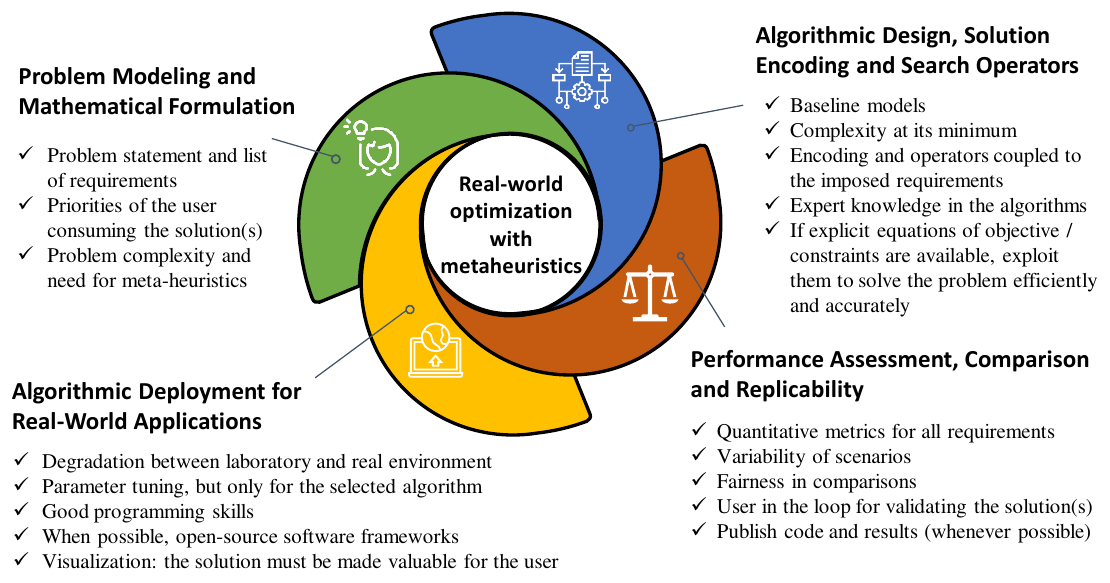}
\caption{Main recommendations given for every phase of our proposed methodology.}
\label{fig:recommendations}
\end{figure}

\begin{enumerate}[leftmargin=*]
\item Problem modeling and mathematical formulation:
\begin{itemize}[leftmargin=*]
\item It should be strictly mandatory to clearly state the problem objectives, variables and constraints, considering all the practical aspects of the scenario at hand (e.g. users consuming the output, contextual factors affecting the validity of the solution, etc.).
\item All non-functional requirements should be exhaustively listed, such as time/memory consumption, accuracy of the solution, the chance to undergo unexpected early interruptions, the usability of the produced solution(s), etc.
\item Objectives and/or functional/non-functional requirements should be prioritized as per the criteria of the user.
\item The complexity of the problem should be analyzed towards substantiating the need for metaheuristics.
\end{itemize}

\item Algorithmic design, solution encoding and search operators:
\begin{itemize}[leftmargin=*]
\item Baseline models should be first searched for in the literature, past experiences, project reports or any other source of information. If they exist, baseline models should be used first:
\begin{itemize}
    \item If any baseline model meets the functional and non-functional requirements, the problem is solved. There is no need for iterating any further.
    \item If no baseline model meets the requirements, they must be considered as a starting point to incrementally improve their compliance with the requirements.
\end{itemize}

\item It is advisable to quantify and trace which requirements benefit the most from each algorithmic modification, so that insighta are gained about which changes can be more promising in order to improve the compliance with every requirement.

\item The complexity of the algorithm must be kept to the minimum required for guaranteeing the requirements, even if the computing technology is capable of running it efficiently. This allows minimizing risks during the deployment of the algorithm.

\item When designing the encoding strategy, population structure and search operators, it is necessary to gauge, when possible, their impact on the degree of fulfillment of the imposed requirements, so that their design becomes coupled to them.

\item Validated algorithmic design templates should be always preferred rather than overly sophisticated algorithmic components. 

\item Expert knowledge acquired over years of observation of the system/asset to be optimized should be always leveraged in the algorithmic design.
\end{itemize}

\item Performance assessment, comparison and replicability:
\begin{itemize}[leftmargin=*]
\item Baseline models selected in the previous phase should be always included in the benchmark.

\item Quantitative metrics must be defined and measured for all functional and non-functional requirements.

\item Variability of scenarios: when the problem at hand can be configured as per a number of parameters, as many problem configurations as possible should be created and evaluated to account for the diversity of scenarios that the algorithm(s) can encounter in practice.

\item For the sake of fairness in the comparisons, parameter tuning must be enforced in all the algorithms of the benchmark (including the baseline ones). Furthermore, statistical tests should be applied to ensure that the gaps among the performance of the algorithms are indeed relevant. 

\item User in the loop: results should be reported comprehensively to ease the decision making process of the end user. It is better to provide several solutions at this phase than in deployment. Furthermore, new requirements often emerge when the user evaluates the results by him/herself. 

\item When soft constraints are considered, the level of constraint fulfillment of the solutions should be also informed to the user.

\item If confidentiality allows it, it is always good and enriching to publish code and results in public repositories. 
\end{itemize}

\item Algorithmic deployment for real-world applications:
\begin{itemize}[leftmargin=*]
\item Parameter tuning of the selected metaheuristic algorithm is a must before proceeding further, so that the eventual performance degradation between the laboratory and the real environment are only due to contextual factors.
\item The degradation of the fulfillment of the requirements when in-lab developments are deployed on the production environment must be quantified and carefully assessed. If needed, a redesign of the algorithm can be enforced to reduce this risk, always departing from the identified cause of the observed degradation.
\item Good programming skills (optimized code, modular, with comments and exception handling) are key for an easy update, extension, and reuse of the developed code for future purposes. 
\item When possible, open-source software frameworks should be selected for the development of the algorithm to be deployed in order to ensure productivity and community support.
\item Hard constraints from corporate development platforms imposed on the implementation language should be taken into account.
\item Straightforward mechanisms to change the parameters of the algorithm should be implemented.
\item Efforts should be conducted towards the visualization of the algorithm's output. How can the solution be made more valuable for the user? Unless a proper answer to this question is given as per the expertise and cognitive profile of the user, this can be a major issue in real-world optimization problems, specially when the user at hand has no technical background whatsoever. 
\end{itemize}
\end{enumerate}

\section{Research Trends in Real-world Optimization with Metaheuristics} \label{sec:prospects}

\begin{figure}[h]
\centering
\includegraphics[width=1.0\columnwidth]{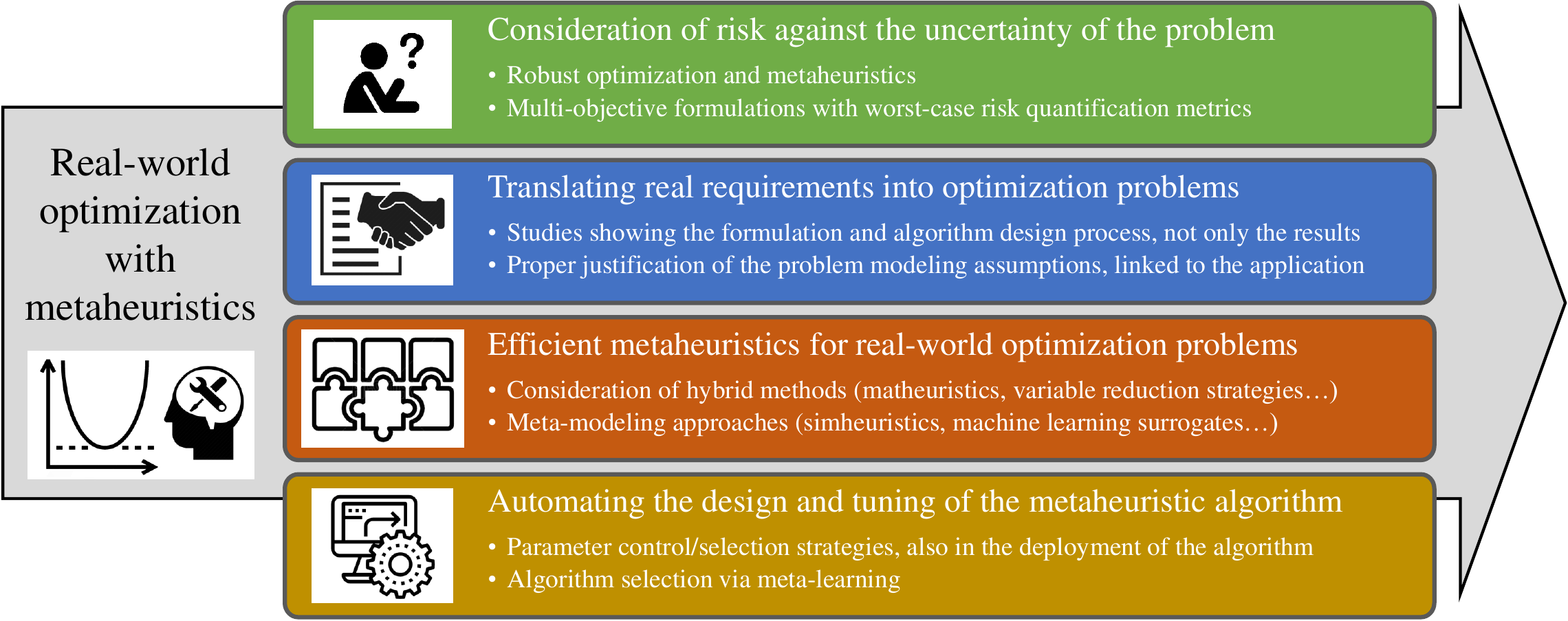}
\caption{Challenges and research directions foreseen for real-world optimization with metaheuristics}
\label{fig:challenges}
\end{figure}

Although the optimization research field has dealt with real-world problems throughout its long life, the diversity and increasing complexity of scenarios in which such problems are formulated in practice have stimulated a plethora of new research directions over the years aimed to manage their different particularities efficiently. In this section, we highlight several challenges and research directions that, given the current state of the art, we consider of utmost relevance for prospective studies in the confluence of real-world optimization and metaheuristics. Our envisioned future for the field is summarized in 
Figure~\ref{fig:challenges}, and elaborated in what follows.

\subsection{Robust optimization and worst-case analysis} \label{ssec:robust}

In real-world optimization scenarios, many sources of uncertainty may arise, from exogenous variables of the environment that cannot be measured and are not considered in the formulation anyhow, to the collected data that can participate in the definition of the objective function(s) and/or constraint(s). Furthermore, it is often the case that, in practice, the user consuming the solution given to the problem is willing to impose worst-case constraints assuming that such sources of uncertainty cannot be counteracted anyhow. In fact, the identification of the worst conditions under which the optimization problem can be formulated is usually much easier for the user than the derivation of efficient strategies to accommodate the uncertainty of the setup. This issue amplifies when tackling the problem at hand with metaheuristics, since the optimization algorithm itself induces an additional source of epistemic uncertainty that may compromise the requisites imposed on the worst case.

This situation unleashes a formidable future for robust optimization, which aims at the design of metaheuristic solvers for problems in which uncertainty is considered explicitly in its formulation \cite{gabrel2014recent}. Initially addressed with tools from mathematical programming, robust optimization has also been studied with metaheuristics, with different approximations to account for uncertainty during the search \cite{jin2005evolutionary,paenke2006efficient}. In this context, a core concept in robust optimization is the level of conservativeness demanded by the user, namely, the level of protection of the solution against the uncertainty of the problem \cite{ben1999robust}. This issue is crucial in real-world optimization, especially for its connection with the notion of risk in circumstances in which decision variables relate to assets that require human intervention. Depending on the implications of implementing the solution in practice, the user might prefer less optimal, albeit safer solutions. For instance, in manufacturing, it is often advisable to be conservative when operating a human-intervened drilling machine. If this operation were to be automated via a metaheuristic algorithm, the solution should ensure a high level of conservativeness with respect to the normal operation of the machine not to engender risks for the operator and/or exposed persons. 

Besides advances reported in this line, we advocate for the analysis of solutions with different quality and levels of conservativeness with respect to the sources of uncertainty existing in the real-world scenario. This analysis can be realized by considering conservativeness as an additional objective function to be minimized, so that multi-objective metaheuristics can be designed to yield an approximation of the Pareto front by considering quality and risk \cite{jin2003trade,deb2009reliability}. When provided with this Pareto front approximation, the user can appraise the implications of the uncertainty on the quality of solutions for the problem, and select the solution with the most practical utility bearing in mind both objectives. We definitely foresee an increasing relevance of risk in real-world optimization problems considering the progressively higher prevalence of automated means to solve them efficiently in real life.

\subsection{Translating real requirements into optimization problems}

A few works have lately revolved around the methodological procedure for formulating an optimization problem, including the definition of its variables and constraints. Assorted tools have been very recently proposed for this purpose, including directed questionnaires \cite{van2020towards} and algebraic modeling languages to describe optimization problems (see \cite{dunning2017jump} and references therein). Despite these tools, there is a large \emph{semantic gap} in practical cases between what the user consuming the solution to the problem truly needs, and what the scientist designing the optimization algorithm understands. Leaving aside non-technicalities that could potentially open this gap, two factors that impact most in widening this gap are 1) the capability of the user and the scientist to \emph{plunge into} the discourse of each other, progressively coming to a point of agreement on \emph{what is needed}; and 2) the capability of the scientist to effectively translate requirements from the application domain into algorithmic clauses. 

The first factor depends roughly on both parties' social and empathetic skills, especially from the scientist who must understand the overall application context in which the problem is framed. A proper understanding of the problem, along with discussions held with the practitioner, can eventually unveil useful insights and hints that help in the problem formulation and the design of the algorithm. For this to occur, the scientist must assimilate all the details concerning the asset/system to be optimized, especially when the objective to be optimized and the imposed constraints cannot be analytically determined.

The second factor promoting the aforementioned semantic gap is more related to the methodology used for the translation between requirements and problem formulation. The issue emerging at this point is whether this process can be enclosed within a unified methodology that comprises all questions and decision steps to be followed for formulating a real-world optimization problem. Unfortunately, the question remains unanswered in the literature, and current practices evince that the definition of a real-world optimization problem is largely ad-hoc and subject to the expertise and modeling skills of the scientist. For instance, in most cases, the number of constraints imposed on a particular problem restricts the search space severely, to the point of modeling it as a constraint satisfaction problem in which the only goal of the solver is to produce a feasible solution. Metaheuristics suitable to deal with constraint satisfaction problems differ from those used for the optimization of an objective function (both single- and multi-objective), as the potential sparsity of the space where feasible solutions are located may call for an extensive use of explorative search operators and diversity-inducing mechanisms. However, there is no clear criterion for shifting in practice to this paradigm. Furthermore, the presence of multiple global optima (the so-called \emph{multi-modality} of the problem's landscape) can be a critical factor for the design of the algorithm. Unless carefully considered from the very inception of the problem, multi-modality can give rise to solutions of no practical use due to non-modeled externalities that discriminate which solutions can be found in practice.

All in all, prospective literature works on real-world optimization should not only restrict their coverage to the presentation of the problem but also design and validate their algorithms. Explanations should also be given on the process by which the formulation of the problem was inferred from the scenario under analysis. In real-world optimization problems, information about the process is almost as valuable as the result itself, inasmuch as the community can largely benefit from innovative methodological practices that can be adopted in other problems.

\subsection{Hybridization of mathematical tools with metaheuristic algorithms}

When addressing real-world optimization problems with metaheuristics, another relevant direction is the hybridization of these algorithms with methods from other disciplines for improved performance of the search process. Such an opportunity arises when the conditions under which the problem is formulated allows for the consideration of additional tools towards enhancing the convergence and/or quality of the solutions elicited by the metaheuristic algorithms. Therefore, the chance to opt for hybrid metaheuristic algorithms is bounded to the case under study and the functional and non-functional requirements imposed thereon, as the incorporation of new search steps in the algorithm might penalize the computational time, increase the memory consumption, entail the purchase of third-party software, or may impose any other similar demand.

One example of this hybridization is the exploitation of explicit formulae defining the objectives and/or constraints. There are plenty of programming methods that can be utilized when the definition of the fitness and constraints comply with certain assumptions, such as a linear or quadratic relationship with the optimization variables. When this is the case, swarm and evolutionary methods for real-world optimization should make use of the aforementioned tools, even if a mathematical formulation of the optimization problem is available. Indeed, if the requirements of the real-world problem under analysis aim at the computational efficiency of the search process, the scientist should do his/her best to benefit from the equations. Unfortunately, this hybridization is not effectively done as per the current state of the art in Evolutionary Algorithms and Swarm Intelligence. Prior work can be found around the exploitation of gradient knowledge of the optimization problem to accelerate local search and ensure feasibility more efficiently in continuous optimization problems \cite{noel2012new}. Domain-specific knowledge is also key for a tailored design of the encoding strategy and other elements of the metaheuristic algorithm \cite{bonissone2006evolutionary}, which in some cases can be inspired by the mathematical foundations of the problem. Search methods capitalizing on the combination of mathematical programming techniques and metaheuristics have been collectively referred to as \emph{matheuristics} \cite{fischetti2018matheuristics}, expanding a flurry of academic contributions in the last years over a series of dedicated workshops. 

In this context, an interesting research path to follow is variable reduction, which can alleviate the computational complexity of the search process by inferring relationships among the system of equations describing a given problem \cite{wu2015variable}. As pointed out in this and other related works, a large gap is still to be bridged to extrapolate these findings to real-world optimization problems lacking properties such as differentiability and continuity. Nevertheless, workarounds can be adopted to infer such relationships and enable variable reduction during the search process, such as approximate means to detect such relationships (via e.g., neural or bayesian networks). Interestingly, reducing part of the variables involved in an optimization problem can bring along an increased complexity of other remaining variables. All this paves the way to integrating variable reduction with traditional mathematical programming methods for constrained optimization, such as the Newton or interior-point methods.

We certainly identify a promising future for the intersection between metaheuristics and traditional mathematical programming methods, especially when solving real-world problems with accurate mathematical equations available. As a matter of fact, several competitions are organized nowadays for the community to share and elaborate on new approaches along this research line. For instance, the competitions on real-world single-objective constrained optimization held at different venues (CEC 2020, SEMCCO 2020, and GECCO 2020) consider a set of 57 real-world constrained problems \cite{kumar2020test}. In these competitions, participants are allowed to use the constraint equations to design the search algorithm. Another example around real-world bound constrained problems can be found in \cite{das2010problem}. In short, we foresee that metaheuristic algorithms hybridized with mathematical programming techniques will become central in future studies related to real-world optimization.

\subsection{Meta-modeling for real-world optimization}

When dealing with physical assets/systems, the evaluation of the quality and/or feasibility of solutions produced over the metaheuristic search can be realized by complex simulation environments mapping the decision variables at their input to the values dictating their fitness/compliance with constraints . The use of digital twins in manufacturing or the design of structures in civil engineering are two recent examples of simulation environments that serve as a computational representation of large-scale complex systems, for which an analytical formulation of all their components and interrelations cannot be easily stated. 

From an optimization perspective, the use of simulators for problem solving (simulation-based optimization or \emph{simheuristics} \cite{juan2015review}) constitutes a straightforward approach to circumvent an issue that appears concurrently in real-world problems: the impossibility of formulating objective functions and constraints in mathematical form. Furthermore, depending on its faithfulness with respect to the modeled asset/system, the use of simheuristics in real-world optimization can also account for the uncertainty present in non-deterministic application scenarios under analysis in a scalable fashion. Consequently, the adoption of simulation-based optimization can ease the quantification of risk incurred by candidate solutions and alternative hypothesis \cite{chica2017simheuristics}, which connects back to our prospects around the importance of risk in real-world optimization (Subsection~\ref{ssec:robust}).

In this context, a research line with a long history in metaheuristic optimization is the use of machine learning surrogates \cite{jin2011surrogate}. Solvers under this paradigm resort to data-based regression techniques to infer the relationship between the decision variables and their objective function value, so that when learned, the evaluation of new candidates for the problem at hand can be efficiently performed by querying the trained regression model \cite{jin2005comprehensive}. Although the alleviation of the computational complexity of the solver is arguably the most extended use of surrogates in metaheuristic optimization, another vein of literature has stressed on the valuable information that surrogates can feed to the search algorithm for improving its convergence. Possibilities for this purpose are diverse, including the evaluation and removal of poor solutions when initializing the population of the metaheuristic algorithm, or the implementation of informed operators that reduce their level of randomness concerning na\"ive metaheuristic implementations \cite{rasheed2000informed}. 

Disregarding the specific model combined with the metaheuristic algorithm (simulation or machine learning surrogates), several problems arise when resorting to these meta-modeling approaches in real-world problems. To begin with, very few works have elaborated on scalable meta-modeling approaches capable of implementing different modeling granularities, each balancing differently between the fidelity of the meta-model with respect to the modeled asset/system, and the computational complexity of the model when queried with a certain input. This trade-off and the challenges stemming therefrom have been widely identified in the related literature \cite{chica2017simheuristics}. We herein underscore the need for further strategies to develop scalable meta-models with varying levels of complexity and fidelity. New advances in this line should marry up with achievements in asynchronous parallel computing, especially when several meta-models are considered jointly, each requiring different complexity levels. This is actually another reason why the prescription of non-functional requirements is of utmost importance in real-world optimization: unless properly accounted from the very beginning, sophisticated meta-models can be of no use if the available computing resources do not fulfill such requirements in practice.

Another issue that remains insufficiently unaddressed to date is how to prevent surrogates from overfitting, specially in problems characterized by many decision variables that are tackled by using complex modeling approaches (e.g., Deep Learning). Under such conditions, and depending on the availability of evaluated examples at the beginning of the search, the learning algorithm might have a few high-dimensional examples available for training the surrogate. This could eventually dominate the learning process and hinder the generalization of the trained model to unseen candidate solutions. Regularization approaches have been extensively suggested to deal with this problem, especially with linear models and neural networks \cite{jin2002framework,bhosekar2018advances}. However, we feel that further research can be pursued towards regularization approaches that, besides overfitting, provide a countermeasure for another serious problem derived from overly complex surrogates: the existence of virtual optima, i.e., optima that do not exist in the original problem under analysis. When this is the case, regularized ensembles and archiving strategies can be effective solutions to both overfitting and virtual optima.

Finally, we briefly pause at the explainability of machine learning models, which currently capitalizes most research contributions reported under the eXplainable Artificial Intelligence (XAI) paradigm \cite{arrieta2020explainable}. XAI refers to all techniques aimed at eliciting interpretable information about the knowledge learned by a model. In the case of black-box surrogate models, XAI must be conceived not only as a driver for acceptability but also as a tool to provide hints for the design of the optimization algorithm. For instance, post-hoc XAI methods can be used to unveil what the surrogate observes in its input (decision variables) to produce a given output (estimated fitness value), so that a global understanding of the decision variables that most correlate with the fitness value can be obtained. This augmented information about the search process can boost the acceptability of the solution provided by the overall surrogate-assisted metaheuristic by a non-expert user. Causal analysis for machine learning models \cite{guo2018survey,moraffah2020causal} can take a step further in this direction, discriminating which decision variables, when modified throughout the search, lead to major changes in the fitness value. These studies can unchain new forms of informed search operators that seize implicit causal relationships between variables and fitness during the search.

\subsection{Automating algorithm selection and parameter tuning}

We round up our prospects with a mention of parameter tuning, which is arguably among the main reasons for differences appearing between the in-lab design of a metaheuristic algorithm and its deployment in a real-world environment. Indeed, the complexity of real-world scenarios can lead to incomplete/oversimplified formulations that do not fully represent the diversity of contextual factors affecting the problem. Furthermore, the problem itself can be dynamic, so that fitness and/or constraints can evolve over time. If this variability is not considered in the definition of the problem nor resolved during the design of the metaheuristic algorithm (by means of e.g., dynamic optimization approaches), differences likely emerge when deploying the metaheuristic in practice. The methodology herein presented contemplates this issue by enforcing a fine-grained tuning of parameters right at the beginning of the application environment, so that the effects of any contextual bias on the compliance of functional and non-functional requirements can be minimized. However, accounting for parameter tuning in our methodology does not play down the fact that parameter tuning is a time-consuming process, especially when the search for a satisfactory parametric configuration of the solver takes into account a mixture of functional and non-functional requirements. 

Fortunately, in this context, the research community has left behind ancient practices in parameter tuning, wherein the metaheuristic algorithms were configured in a trial-and-error fashion, or by using the configurations utilized in similar studies. This procedure is by no means acceptable in academic works comparing among metaheuristics, nor should this be the case in real-world optimization. Vigorous research is nowadays concentrated on the derivation of new algorithmic means to automate the process of adjusting the parameters of metaheuristic solvers, either during the search process (self-adaptation mechanisms for parameter \emph{control}) or as a separate off-line process performed before the configured metaheuristic is actually executed (parameter \emph{tuning}). Both approaches can actually be applied to real-world problems, yet the recent activity noted in the field is steering more notably towards parameter tuning approaches due to their independence with respect to the metaheuristic algorithm to be adjusted. In any case, when used in real-world optimization, automated parameter tuning methods can not only ease this process for non-expert users, but also perform it more efficiently than grid search methods.

Automated parameter tuning has so far provided a rich substrate of methods and software frameworks, mature enough for their early adoption to cope with real-world problems of realistic complexity \cite{huang2020survey}. Our claim in this matter is that the flexibility of current automated parameter tuning frameworks is limited, and leaves aside non-functional requirements that often emerge in real-world environments. Most of them focus on optimality, i.e., on finding a configuration of the metaheuristic algorithm that performs best as per the objective function(s) of the problem at hand. In many situations, this goal suffices for the interest of the user. However, we utterly believe that other aspects should also be reflected in this process, such as implementation complexity (time, memory), simplicity of search operators, and robustness of the configured algorithm against factors inducing uncertainty in the definition of the problem. All in all, functional and non-functional requirements of real-world problems are also affected by the parametric configuration of metaheuristics, so there is a pressing need for embedding metrics that quantify such non-functional requirements in existing frameworks.

Finally, we dedicate some closing words to the field of meta-learning, which is understood as the family of methods aimed at inferring a potentially good algorithm for a given problem without actually addressing it, namely, just by the similarities of the problem with others tackled in the past \cite{smith2008towards,kotthoff2016algorithm}. For this purpose, meta-learning methods for optimization problems usually hinge on the extraction of meta-features from the given problem, which are then used as inputs of a supervised learning model that recommends the best algorithm \cite{smith2011discovering}. In other words, meta-learning approaches automate the same task than that of automated parameter tuning methods, but without the computational complexity required by the latter to evaluate multiple candidate solutions representing the algorithm and/or its parameter values. Studies on meta-learning for the recommendation of metaheuristic algorithms have so far been centered on instances of a few classical optimization problems (e.g., traveling salesman \cite{kanda2016meta}, vehicle routing \cite{gutierrez2019selecting} or flow-shop scheduling \cite{pavelski2019meta}). In those works meta-features are extracted from graph representations of the problem under analysis, or the analysis of their fitness landscapes. However, such meta-features are largely problem-dependent, which leaves an open question on whether such meta-features can attain a good generalization performance of the meta-learner when facing real-world problems, in which problem formulations can be much more diverse in practice. Furthermore, the discovery of alternative meta-feature extraction methods can pave the way to the consideration of meta-learning methods as a first step towards the automated construction of optimization ensembles, which are known to be less sensitive to the parametric configuration of their constituent solvers than single metaheuristics. Moreover, ensembles of methods can be also used to identify which operator, parameter value or algorithmic component is effective for a particular problem in a competitive way, and just in one step \cite{wu2019ensemble}. We see a fascinating opportunity for meta-learning in real-world optimization with metaheuristics, sparking many research directions for achieving higher degrees of intelligent design automation as the ones reviewed heretofore.

\section{Conclusions and Outlook} \label{sec:conc}

In this tutorial, we have proposed an end-to-end methodology for addressing real-world optimization problems with metaheuristic algorithms. Our methodology covers from the identification of the optimization problem itself to the deployment of the metaheuristic algorithm, including the determination of functional and non-functional requirements, the design of the metaheuristic itself, validation, and benchmarking. Each step comprising our methodology has been explained in detail along with an enumeration of the technical aspects that should be considered by both the scientist designing the algorithm and the user consuming its output. Recommendations are also given for newcomers to avoid misconceptions and bad practices observed in the literature related to real-world optimization.

We have complemented our prescribed methodology with a set of challenges and research directions which, according to our experience and assessment of the current status of the field, should drive efforts in years to come. Specifically, our vision gravitates around four different domains:
\begin{itemize}[leftmargin=*]
    \item The consideration of risk as an additional objective to be minimized, and the massive adoption of robust optimization techniques, given the high uncertainty under which real-world optimization problems are formulated and the inherent stochastic nature of metaheuristic algorithms.
    \item More reported evidences of the process by which real-world optimization problems are addressed, expanding the scientific value of prospective studies not only to the algorithm(s) and provided solution(s), but also to the inception of the problem and the storytelling themselves.
    \item There is a need for efficient means to cope with the complexity of real-world problems during the metaheuristic search, in which we claim that the hybridization with mathematical tools, meta-modeling, and machine learning surrogates will have an increasingly prominent role in the field.
    \item The incorporation of intelligent methods to automate the selection and parameter tuning of the metaheuristic algorithm, which requires current automated parameter tuning frameworks and meta-learning approaches to consider metrics related to functional and non-functional requirements imposed in real-world scenarios.
\end{itemize}

We hope that the methodology proposed in this article and our prospects serve as a guiding light for upcoming research works falling in the confluence between metaheuristic algorithms and real-world optimization. It is our firm belief that the inherent complexity and uncertainty of real-world problems has to be boarded with the methodological rigor required to ensure the practical value of the developed metaheuristics. Unless common methodological standards for real-world optimization are embraced in the future, a major gap will remain unbridged between academia, industrial stakeholders, and the society as a whole.

\section*{Acknowledgements}

Eneko Osaba, Esther Villar-Rodriguez and Javier Del Ser would like to thank the Basque Government through EMAITEK and ELKARTEK (ref. 3KIA) funding grants. Javier Del Ser also acknowledges funding support from the Department of Education of the Basque Government (Consolidated Research Group MATHMODE, IT1294-19). Antonio LaTorre acknowledges funding from the Spanish Ministry of Science (TIN2017-83132-C2-2-R). Carlos A. Coello Coello acknowledges support from CONACyT grant no. 2016-01-1920 ({\em Investigaci\'{o}n en Fronteras de la Ciencia 2016}) and from a SEP-Cinvestav grant (application no. 4). Francisco Herrera and Daniel Molina are partially supported by the project DeepSCOP-Ayudas Fundaci\'on BBVA a Equipos de Investigaci\'on Cient\'ifica en Big Data 2018, and the Spanish Ministry of Science and Technology under  project TIN2017-89517-P.  

\bibliography{mybibfile,refs,atorre,dmolina}

\begin{thebibliography}{100}
\expandafter\ifx\csname url\endcsname\relax
  \def\url#1{\texttt{#1}}\fi
\expandafter\ifx\csname urlprefix\endcsname\relax\def\urlprefix{URL }\fi
\expandafter\ifx\csname href\endcsname\relax
  \def\href#1#2{#2} \def\path#1{#1}\fi

\bibitem{hussain2019metaheuristic}
K.~Hussain, M.~N.~M. Salleh, S.~Cheng, Y.~Shi, Metaheuristic research: a
  comprehensive survey, Artificial Intelligence Review 52~(4) (2019)
  2191--2233.

\bibitem{boussaid2013survey}
I.~Boussa{\"\i}D, J.~Lepagnot, P.~Siarry, A survey on optimization
  metaheuristics, Information sciences 237 (2013) 82--117.

\bibitem{kennedy2006swarm}
J.~Kennedy, Swarm intelligence, in: Handbook of nature-inspired and innovative
  computing, Springer, 2006, pp. 187--219.

\bibitem{eiben2015evolutionary}
A.~E. Eiben, J.~Smith, From evolutionary computation to the evolution of
  things, Nature 521~(7553) (2015) 476--482.

\bibitem{del2019bio}
J.~Del~Ser, E.~Osaba, D.~Molina, X.-S. Yang, S.~Salcedo-Sanz, D.~Camacho,
  S.~Das, P.~N. Suganthan, C.~A.~C. Coello, F.~Herrera, Bio-inspired
  computation: Where we stand and what's next, Swarm and Evolutionary
  Computation 48 (2019) 220--250.

\bibitem{yang2018mathematical}
X.-S. Yang, Mathematical analysis of nature-inspired algorithms, in:
  Nature-Inspired Algorithms and Applied Optimization, Springer, 2018, pp.
  1--25.

\bibitem{pranzo2016iterated}
M.~Pranzo, D.~Pacciarelli, An iterated greedy metaheuristic for the blocking
  job shop scheduling problem, Journal of Heuristics 22~(4) (2016) 587--611.

\bibitem{vidal2015hybrid}
T.~Vidal, M.~Battarra, A.~Subramanian, G.~Erdogan, Hybrid metaheuristics for
  the clustered vehicle routing problem, Computers \& Operations Research 58
  (2015) 87--99.

\bibitem{danka2013statistically}
S.~Danka, A statistically correct methodology to compare metaheuristics in
  resource-constrained project scheduling, Pollack Periodica 8~(3) (2013)
  119--126.

\bibitem{kendall2016good}
G.~Kendall, R.~Bai, J.~B{\l}azewicz, P.~De~Causmaecker, M.~Gendreau, R.~John,
  J.~Li, B.~McCollum, E.~Pesch, R.~Qu, et~al., Good laboratory practice for
  optimization research, Journal of the Operational Research Society 67~(4)
  (2016) 676--689.

\bibitem{jaszkiewicz2004evaluation}
A.~Jaszkiewicz, Evaluation of multiple objective metaheuristics, in:
  Metaheuristics for multiobjective optimisation, Springer, 2004, pp. 65--89.

\bibitem{chiarandini2007experiments}
M.~Chiarandini, L.~Paquete, M.~Preuss, E.~Ridge, Experiments on metaheuristics:
  Methodological overview and open issues, Tech. rep., Technical Report
  DMF-2007-03-003, The Danish Mathematical Society, Denmark (2007).

\bibitem{hochba1997approximation}
D.~S. Hochba, Approximation algorithms for np-hard problems, ACM Sigact News
  28~(2) (1997) 40--52.

\bibitem{papadimitriou1998combinatorial}
C.~H. Papadimitriou, K.~Steiglitz, Combinatorial optimization: algorithms and
  complexity, Courier Corporation, 1998.

\bibitem{papadimitriou1991optimization}
C.~H. Papadimitriou, M.~Yannakakis, Optimization, approximation, and complexity
  classes, Journal of computer and system sciences 43~(3) (1991) 425--440.

\bibitem{blum2003metaheuristics}
C.~Blum, A.~Roli, Metaheuristics in combinatorial optimization: Overview and
  conceptual comparison, ACM computing surveys (CSUR) 35~(3) (2003) 268--308.

\bibitem{dokeroglu2019survey}
T.~Dokeroglu, E.~Sevinc, T.~Kucukyilmaz, A.~Cosar, A survey on new generation
  metaheuristic algorithms, Computers \& Industrial Engineering (2019) 106040.

\bibitem{yang2013swarm}
X.-S. Yang, Z.~Cui, R.~Xiao, A.~H. Gandomi, M.~Karamanoglu, Swarm intelligence
  and bio-inspired computation: theory and applications, Newnes, 2013.

\bibitem{bonabeau1999swarm}
E.~Bonabeau, D.~d. R. D.~F. Marco, M.~Dorigo, G.~Theraulaz, et~al., Swarm
  intelligence: from natural to artificial systems, Oxford university press,
  1999.

\bibitem{de2006evolutionary}
K.~A. De~Jong, Evolutionary computation: a unified approach, MIT press, 2006.

\bibitem{genetic1}
D.~Goldberg, Genetic algorithms in search, optimization, and machine learning,
  Addison-Wesley Professional, 1989.

\bibitem{genetic2}
K.~De~Jong, Analysis of the behavior of a class of genetic adaptive systems,
  Ph.D. thesis, University of Michigan, Michigan, USA (1975).

\bibitem{PSO}
J.~Kennedy, R.~Eberhart, et~al., Particle swarm optimization, in: Proceedings
  of IEEE international conference on neural networks, Vol.~4, Perth,
  Australia, 1995, pp. 1942--1948.

\bibitem{dorigo1997ant}
M.~Dorigo, L.~M. Gambardella, Ant colony system: a cooperative learning
  approach to the traveling salesman problem, IEEE Transactions on evolutionary
  computation 1~(1) (1997) 53--66.

\bibitem{molina2020comprehensive}
D.~Molina, J.~Poyatos, J.~Del~Ser, S.~Garc{\'\i}a, A.~Hussain, F.~Herrera,
  Comprehensive taxonomies of nature-and bio-inspired optimization: Inspiration
  versus algorithmic behavior, critical analysis and recommendations, arXiv
  preprint arXiv:2002.08136 (2020).

\bibitem{sorensen2015metaheuristics}
K.~S{\"o}rensen, Metaheuristics—the metaphor exposed, International
  Transactions in Operational Research 22~(1) (2015) 3--18.

\bibitem{sorensen2018history}
K.~S{\"o}rensen, M.~Sevaux, F.~Glover, A history of metaheuristics, Handbook of
  heuristics (2018) 1--18.

\bibitem{derrac2011practical}
J.~Derrac, S.~Garc{\'\i}a, D.~Molina, F.~Herrera, A practical tutorial on the
  use of nonparametric statistical tests as a methodology for comparing
  evolutionary and swarm intelligence algorithms, Swarm and Evolutionary
  Computation 1~(1) (2011) 3--18.

\bibitem{braysy2005vehicle}
O.~Br{\"a}ysy, M.~Gendreau, Vehicle routing problem with time windows, part i:
  Route construction and local search algorithms, Transportation science 39~(1)
  (2005) 104--118.

\bibitem{osaba2018good}
E.~Osaba, R.~Carballedo, F.~Diaz, E.~Onieva, A.~D. Masegosa, A.~Perallos, Good
  practice proposal for the implementation, presentation, and comparison of
  metaheuristics for solving routing problems, Neurocomputing 271 (2018) 2--8.

\bibitem{eggensperger2019pitfalls}
K.~Eggensperger, M.~Lindauer, F.~Hutter, Pitfalls and best practices in
  algorithm configuration, Journal of Artificial Intelligence Research 64
  (2019) 861--893.

\bibitem{eiben2002critical}
A.~E. Eiben, M.~Jelasity, A critical note on experimental research methodology
  in ec, in: Proceedings of the 2002 Congress on Evolutionary Computation.
  CEC'02 (Cat. No. 02TH8600), Vol.~1, IEEE, 2002, pp. 582--587.

\bibitem{vcrepinvsek2014replication}
M.~{\v{C}}repin{\v{s}}ek, S.-H. Liu, M.~Mernik, Replication and comparison of
  computational experiments in applied evolutionary computing: common pitfalls
  and guidelines to avoid them, Applied Soft Computing 19 (2014) 161--170.

\bibitem{latorre2020fairness}
A.~LaTorre, D.~Molina, E.~Osaba, J.~Del~Ser, F.~Herrera, Fairness in
  bio-inspired optimization research: A prescription of methodological
  guidelines for comparing meta-heuristics, arXiv preprint arXiv:2004.09969
  (2020).

\bibitem{hansen2016coco}
N.~Hansen, A.~Auger, D.~Brockhoff, D.~Tu{\v{s}}ar, T.~Tu{\v{s}}ar, Coco:
  Performance assessment, arXiv preprint arXiv:1605.03560 (2016).

\bibitem{edmonds_2008}
J.~Edmonds, Definition of Optimization Problems, Cambridge University Press,
  2008, Ch.~13, p. 171–172.
\newblock \href {https://doi.org/10.1017/CBO9780511808241.015}
  {\path{doi:10.1017/CBO9780511808241.015}}.

\bibitem{huang2007problem}
V.~Huang, A.~K. Qin, K.~Deb, E.~Zitzler, P.~N. Suganthan, J.~Liang, M.~Preuss,
  S.~Huband, Problem definitions for performance assessment of multi-objective
  optimization algorithms, Tech. rep., School of EEE, Nanyang Technological
  University (2007).

\bibitem{kumar2019research}
R.~Kumar, Research methodology: A step-by-step guide for beginners, Sage
  Publications Limited, 2010.

\bibitem{jie2019two}
W.~Jie, J.~Yang, M.~Zhang, Y.~Huang, The two-echelon capacitated electric
  vehicle routing problem with battery swapping stations: Formulation and
  efficient methodology, European Journal of Operational Research 272~(3)
  (2019) 879--904.

\bibitem{delorme2016bin}
M.~Delorme, M.~Iori, S.~Martello, Bin packing and cutting stock problems:
  Mathematical models and exact algorithms, European Journal of Operational
  Research 255~(1) (2016) 1--20.

\bibitem{glinz2007non}
M.~Glinz, On non-functional requirements, in: 15th IEEE International
  Requirements Engineering Conference (RE 2007), IEEE, 2007, pp. 21--26.

\bibitem{robertson2012mastering}
S.~Robertson, J.~Robertson, Mastering the requirements process: Getting
  requirements right, Addison-wesley, 2012.

\bibitem{sommerville2001software}
I.~Sommerville, Software engineering, Ed., Harlow, UK.: Addison-Wesley (2001).

\bibitem{davis1993software}
M.~Davis, Software requirements, OBJECTS FUNCTIONS \& STATUS (1993).

\bibitem{approx1996}
E.~Coffman, M.~Garey, D.~Johnson, Approximation Algorithms for NP-Hard
  Problems, 1996, pp. 46--93.

\bibitem{lange2000optimization}
K.~Lange, D.~R. Hunter, I.~Yang, Optimization transfer using surrogate
  objective functions, Journal of computational and graphical statistics 9~(1)
  (2000) 1--20.

\bibitem{Spagnol2019}
A.~Spagnol, R.~L. Riche, S.~D. Veiga,
  \href{http://dx.doi.org/10.1137/18M1167978}{Global sensitivity analysis for
  optimization with variable selection}, SIAM/ASA Journal on Uncertainty
  Quantification 7~(2) (2019) 417–443.
\newblock \href {https://doi.org/10.1137/18m1167978}
  {\path{doi:10.1137/18m1167978}}.
\newline\urlprefix\url{http://dx.doi.org/10.1137/18M1167978}

\bibitem{boyd2004convex}
S.~Boyd, S.~P. Boyd, L.~Vandenberghe, Convex optimization, Cambridge university
  press, 2004.

\bibitem{ponton2017}
B.~{Ponton}, A.~{Herzog}, A.~{Del Prete}, S.~{Schaal}, L.~{Righetti}, On time
  optimization of centroidal momentum dynamics, in: 2018 IEEE International
  Conference on Robotics and Automation (ICRA), 2018, pp. 5776--5782.

\bibitem{wright1932roles}
S.~Wright, The roles of mutation, inbreeding, crossbreeding, and selection in
  evolution, Vol.~1, na, 1932.

\bibitem{reidys2002combinatorial}
C.~M. Reidys, P.~F. Stadler, Combinatorial landscapes, SIAM review 44~(1)
  (2002) 3--54.

\bibitem{pitzer2012comprehensive}
E.~Pitzer, M.~Affenzeller, A comprehensive survey on fitness landscape
  analysis, in: Recent advances in intelligent engineering systems, Springer,
  2012, pp. 161--191.

\bibitem{merz1999fitness}
P.~Merz, B.~Freisleben, et~al., Fitness landscapes and memetic algorithm
  design, New ideas in optimization (1999) 245--260.

\bibitem{ronald1997robust}
S.~Ronald, Robust encodings in genetic algorithms: A survey of encoding issues,
  in: Proceedings of 1997 IEEE International Conference on Evolutionary
  Computation (ICEC'97), IEEE, 1997, pp. 43--48.

\bibitem{talbi2009metaheuristics}
E.-G. Talbi, Metaheuristics: from design to implementation, Vol.~74, John Wiley
  \& Sons, 2009.

\bibitem{chakraborty2003analysis}
U.~K. Chakraborty, C.~Z. Janikow, An analysis of gray versus binary encoding in
  genetic search, Information Sciences 156~(3-4) (2003) 253--269.

\bibitem{bierwirth1996permutation}
C.~Bierwirth, D.~C. Mattfeld, H.~Kopfer, On permutation representations for
  scheduling problems, in: International Conference on Parallel Problem Solving
  from Nature, Springer, 1996, pp. 310--318.

\bibitem{bean1994genetic}
J.~C. Bean, Genetic algorithms and random keys for sequencing and optimization,
  ORSA journal on computing 6~(2) (1994) 154--160.

\bibitem{rothlauf2006representations}
F.~Rothlauf, Representations for genetic and evolutionary algorithms, in:
  Representations for Genetic and Evolutionary Algorithms, Springer, 2006, pp.
  9--32.

\bibitem{larranaga1999genetic}
P.~Larranaga, C.~M.~H. Kuijpers, R.~H. Murga, I.~Inza, S.~Dizdarevic, Genetic
  algorithms for the travelling salesman problem: A review of representations
  and operators, Artificial Intelligence Review 13~(2) (1999) 129--170.

\bibitem{dorigo1999ant}
M.~Dorigo, G.~Di~Caro, Ant colony optimization: a new meta-heuristic, in:
  Proceedings of the 1999 congress on evolutionary computation-CEC99 (Cat. No.
  99TH8406), Vol.~2, IEEE, 1999, pp. 1470--1477.

\bibitem{blum2002ant}
C.~Blum, M.~Sampels, Ant colony optimization for fop shop scheduling: a case
  study on different pheromone representations, in: Proceedings of the 2002
  Congress on Evolutionary Computation. CEC'02 (Cat. No. 02TH8600), Vol.~2,
  IEEE, 2002, pp. 1558--1563.

\bibitem{osaba2018multi}
E.~Osaba, J.~Del~Ser, A.~J. Nebro, I.~La{\~n}a, M.~N. Bilbao, J.~J.
  Sanchez-Medina, Multi-objective optimization of bike routes for last-mile
  package delivery with drop-offs, in: 2018 21st International Conference on
  Intelligent Transportation Systems (ITSC), IEEE, 2018, pp. 865--870.

\bibitem{salcedo2002feature}
S.~Salcedo-Sanz, M.~Prado-Cumplido, F.~P{\'e}rez-Cruz,
  C.~Bouso{\~n}o-Calz{\'o}n, Feature selection via genetic optimization, in:
  International Conference on Artificial Neural Networks, Springer, 2002, pp.
  547--552.

\bibitem{salcedo2004enhancing}
S.~Salcedo-Sanz, G.~Camps-Valls, F.~P{\'e}rez-Cruz, J.~Sep{\'u}lveda-Sanchis,
  C.~Bouso{\~n}o-Calz{\'o}n, Enhancing genetic feature selection through
  restricted search and walsh analysis, IEEE Transactions on Systems, Man, and
  Cybernetics, Part C (Applications and Reviews) 34~(4) (2004) 398--406.

\bibitem{salcedo2007improving}
S.~Salcedo-Sanz, J.~Su, Improving metaheuristics convergence properties in
  inductive query by example using two strategies for reducing the search
  space, Computers \& operations research 34~(1) (2007) 91--106.

\bibitem{gupta2015multifactorial}
A.~Gupta, Y.-S. Ong, L.~Feng, Multifactorial evolution: toward evolutionary
  multitasking, IEEE Transactions on Evolutionary Computation 20~(3) (2015)
  343--357.

\bibitem{gupta2017insights}
A.~Gupta, Y.-S. Ong, L.~Feng, Insights on transfer optimization: Because
  experience is the best teacher, IEEE Transactions on Emerging Topics in
  Computational Intelligence 2~(1) (2017) 51--64.

\bibitem{kirkpatrick1983optimization}
S.~Kirkpatrick, C.~D. Gelatt, M.~P. Vecchi, Optimization by simulated
  annealing, science 220~(4598) (1983) 671--680.

\bibitem{glover1998tabu}
F.~Glover, M.~Laguna, Tabu search, in: Handbook of combinatorial optimization,
  Springer, 1998, pp. 2093--2229.

\bibitem{alba2005parallel}
E.~Alba, Parallel metaheuristics: a new class of algorithms, Vol.~47, John
  Wiley \& Sons, 2005.

\bibitem{alba2013parallel}
E.~Alba, G.~Luque, S.~Nesmachnow, Parallel metaheuristics: recent advances and
  new trends, International Transactions in Operational Research 20~(1) (2013)
  1--48.

\bibitem{atashpaz2007imperialist}
E.~Atashpaz-Gargari, C.~Lucas, Imperialist competitive algorithm: an algorithm
  for optimization inspired by imperialistic competition, in: 2007 IEEE
  congress on evolutionary computation, IEEE, 2007, pp. 4661--4667.

\bibitem{luque2011parallel}
G.~Luque, E.~Alba, Parallel genetic algorithms: theory and real world
  applications, Vol. 367, Springer, 2011.

\bibitem{cantu1998survey}
E.~Cant{\'u}-Paz, A survey of parallel genetic algorithms, Calculateurs
  paralleles, reseaux et systems repartis 10~(2) (1998) 141--171.

\bibitem{karaboga2007artificial}
D.~Karaboga, B.~Basturk, Artificial bee colony (abc) optimization algorithm for
  solving constrained optimization problems, in: International fuzzy systems
  association world congress, Springer, 2007, pp. 789--798.

\bibitem{yang2009cuckoo}
X.-S. Yang, S.~Deb, Cuckoo search via l{\'e}vy flights, in: 2009 World Congress
  on Nature \& Biologically Inspired Computing (NaBIC), IEEE, 2009, pp.
  210--214.

\bibitem{das2011real}
S.~Das, S.~Maity, B.-Y. Qu, P.~N. Suganthan, Real-parameter evolutionary
  multimodal optimization—a survey of the state-of-the-art, Swarm and
  Evolutionary Computation 1~(2) (2011) 71--88.

\bibitem{yang2009firefly}
X.-S. Yang, Firefly algorithms for multimodal optimization, in: International
  symposium on stochastic algorithms, Springer, 2009, pp. 169--178.

\bibitem{sivaraj2011review}
R.~Sivaraj, T.~Ravichandran, A review of selection methods in genetic
  algorithm, International journal of engineering science and technology 3~(5)
  (2011) 3792--3797.

\bibitem{prakasam2016metaheuristic}
A.~Prakasam, N.~Savarimuthu, Metaheuristic algorithms and probabilistic
  behaviour: a comprehensive analysis of ant colony optimization and its
  variants, Artificial Intelligence Review 45~(1) (2016) 97--130.

\bibitem{olafsson2006metaheuristics}
S.~{\'O}lafsson, Metaheuristics, Handbooks in operations research and
  management science 13 (2006) 633--654.

\bibitem{kerschke2019automated}
P.~Kerschke, H.~H. Hoos, F.~Neumann, H.~Trautmann, Automated algorithm
  selection: Survey and perspectives, Evolutionary computation 27~(1) (2019)
  3--45.

\bibitem{wolpert1995no}
D.~H. Wolpert, W.~G. Macready, et~al., No free lunch theorems for search, Tech.
  rep., Technical Report SFI-TR-95-02-010, Santa Fe Institute (1995).

\bibitem{iacca2012ockham}
G.~Iacca, F.~Neri, E.~Mininno, Y.-S. Ong, M.-H. Lim, Ockham’s razor in
  memetic computing: three stage optimal memetic exploration, Information
  Sciences 188 (2012) 17--43.

\bibitem{caraffini2012three}
F.~Caraffini, G.~Iacca, F.~Neri, E.~Mininno, Three variants of three stage
  optimal memetic exploration for handling non-separable fitness landscapes,
  in: 2012 12th UK Workshop on Computational Intelligence (UKCI), IEEE, 2012,
  pp. 1--8.

\bibitem{cotta2008adaptive}
C.~Cotta, M.~Sevaux, K.~S{\"o}rensen, Adaptive and multilevel metaheuristics,
  Vol. 136, Springer, 2008.

\bibitem{woodward2011automatically}
J.~R. Woodward, J.~Swan, Automatically designing selection heuristics, in:
  Proceedings of the 13th annual conference companion on Genetic and
  evolutionary computation, 2011, pp. 583--590.

\bibitem{woodward2012automatic}
J.~R. Woodward, J.~Swan, The automatic generation of mutation operators for
  genetic algorithms, in: Proceedings of the 14th annual conference companion
  on Genetic and evolutionary computation, 2012, pp. 67--74.

\bibitem{liuParadoxesNumericalComparison2020}
Q.~Liu, W.~V. Gehrlein, L.~Wang, Y.~Yan, Y.~Cao, W.~Chen, Y.~Li, Paradoxes in
  {{Numerical Comparison}} of {{Optimization Algorithms}}, IEEE Transactions on
  Evolutionary Computation 24~(4) (2020) 777--791.
\newblock \href {https://doi.org/10.1109/TEVC.2019.2955110}
  {\path{doi:10.1109/TEVC.2019.2955110}}.

\bibitem{tanabe2020easy}
R.~Tanabe, H.~Ishibuchi, An easy-to-use real-world multi-objective optimization
  problem suite, Applied Soft Computing 89 (2020) 106078.

\bibitem{cheng2017benchmark}
R.~Cheng, M.~Li, Y.~Tian, X.~Zhang, S.~Yang, Y.~Jin, X.~Yao, A benchmark test
  suite for evolutionary many-objective optimization, Complex \& Intelligent
  Systems 3~(1) (2017) 67--81.

\bibitem{chen2020proposal}
W.~Chen, H.~Ishibuchi, K.~Shang, Proposal of a realistic many-objective test
  suite, in: International Conference on Parallel Problem Solving from Nature,
  Springer, 2020, pp. 201--214.

\bibitem{picard2020realistic}
C.~Picard, J.~Schiffmann, Realistic constrained multi-objective optimization
  benchmark problems from design, IEEE Transactions on Evolutionary Computation
  (2020).

\bibitem{kumar2020test}
A.~Kumar, G.~Wu, M.~Z. Ali, R.~Mallipeddi, P.~N. Suganthan, S.~Das, A
  test-suite of non-convex constrained optimization problems from the
  real-world and some baseline results, Swarm and Evolutionary Computation
  (2020) 100693.

\bibitem{he2019repository}
C.~He, Y.~Tian, H.~Wang, Y.~Jin, A repository of real-world datasets for
  data-driven evolutionary multiobjective optimization, Complex \& Intelligent
  Systems (2019) 1--9.

\bibitem{lou2019constructing}
Y.~Lou, S.~Y. Yuen, On constructing alternative benchmark suite for
  evolutionary algorithms, Swarm and evolutionary computation 44 (2019)
  287--292.

\bibitem{ishibuchi2019scalable}
H.~Ishibuchi, Y.~Peng, K.~Shang, A scalable multimodal multiobjective test
  problem, in: 2019 IEEE Congress on Evolutionary Computation (CEC), IEEE,
  2019, pp. 310--317.

\bibitem{omidvar2015designing}
M.~N. Omidvar, X.~Li, K.~Tang, Designing benchmark problems for large-scale
  continuous optimization, Information Sciences 316 (2015) 419--436.

\bibitem{moreBenchmarkingDerivativeFreeOptimization2009}
J.~J. Mor{\'e}, S.~M. Wild, Benchmarking {{Derivative}}-{{Free Optimization
  Algorithms}}, SIAM Journal on Optimization 20~(1) (2009) 172--191.
\newblock \href {https://doi.org/10.1137/080724083}
  {\path{doi:10.1137/080724083}}.

\bibitem{liuBenchmarkingStochasticAlgorithms2017}
Q.~Liu, W.-N. Chen, J.~D. Deng, T.~Gu, H.~Zhang, Z.~Yu, J.~Zhang, Benchmarking
  {{Stochastic Algorithms}} for {{Global Optimization Problems}} by
  {{Visualizing Confidence Intervals}}, IEEE Transactions on Cybernetics 47~(9)
  (2017) 2924--2937.
\newblock \href {https://doi.org/10.1109/TCYB.2017.2659659}
  {\path{doi:10.1109/TCYB.2017.2659659}}.

\bibitem{LaTorre2010d}
A.~LaTorre, S.~Muelas, J.~M. Pe{\~n}a, A {{MOS}}-based dynamic memetic
  differential evolution algorithm for continuous optimization: A scalability
  test, Soft Computing - A Fusion of Foundations, Methodologies and
  Applications 15~(11) (2010) 2187--2199.
\newblock \href {https://doi.org/10.1007/s00500-010-0646-3}
  {\path{doi:10.1007/s00500-010-0646-3}}.

\bibitem{Herrera-Poyatos2017}
A.~{Herrera-Poyatos}, F.~Herrera, Genetic and {{Memetic Algorithm}} with
  {{Diversity Equilibrium}} based on {{Greedy Diversification}}, CoRR
  abs/1702.03594 (2017).

\bibitem{Crepinsek2013}
M.~{\v C}repin{\v s}ek, S.~H. Liu, M.~Mernik, Exploration and {{Exploitation}}
  in {{Evolutionary Algorithms}}: A {{Survey}}, ACM Computing Surveys 45~(3)
  (2013) 1--33.
\newblock \href {https://doi.org/10.1145/2480741.2480752}
  {\path{doi:10.1145/2480741.2480752}}.

\bibitem{McCabe1976}
T.~J. McCabe, {A Complexity Measure}, IEEE Transactions on Software Engineering
  SE-2~(4) (1976) 308--320.

\bibitem{demsarStatisticalComparisonsClassifiers2006}
J.~Dem{\v s}ar, Statistical {{Comparisons}} of {{Classifiers}} over {{Multiple
  Data Sets}}, The Journal of Machine Learning Research 7 (2006) 1--30.

\bibitem{greenlandStatisticalTestsValues2016}
S.~Greenland, S.~Senn, K.~Rothman, J.~Carlin, C.~Poole, S.~Goodman, D.~Altman,
  Statistical tests, {{P}} values, confidence intervals, and power: A guide to
  misinterpretations, European Journal of Epidemiology 31~(4) (2016) 337--350.
\newblock \href {https://doi.org/10.1007/s10654-016-0149-3}
  {\path{doi:10.1007/s10654-016-0149-3}}.

\bibitem{bayes}
A.~Benavoli, G.~Corani, J.~Dem\v{s}ar, M.~Zaffalon, Time for a change: A
  tutorial for comparing multiple classifiers through bayesian analysis, The
  Journal of Machine Learning Research 18~(1) (2017) 2653–2688.

\bibitem{biedrzyckiEquivalenceAlgorithmImplementations2019}
R.~Biedrzycki, On equivalence of algorithm's implementations: The
  {{CMA}}-{{ES}} algorithm and its five implementations, in: Proceedings of the
  {{Genetic}} and {{Evolutionary Computation Conference Companion}}, {{GECCO}}
  '19, {Association for Computing Machinery}, {Prague, Czech Republic}, 2019,
  pp. 247--248.
\newblock \href {https://doi.org/10.1145/3319619.3322011}
  {\path{doi:10.1145/3319619.3322011}}.

\bibitem{killeenPredictControlReplicate2019}
P.~Killeen, Predict, {{Control}}, and {{Replicate}} to {{Understand}}: {{How
  Statistics Can Foster}} the {{Fundamental Goals}} of {{Science}},
  Perspectives on Behavior Science 42~(1) (2019) 109--132.
\newblock \href {https://doi.org/10.1007/s40614-018-0171-8}
  {\path{doi:10.1007/s40614-018-0171-8}}.

\bibitem{pengReproducibleResearchComputational2011}
R.~D. Peng, Reproducible {{Research}} in {{Computational Science}}, Science
  334~(6060) (2011) 1226--1227.
\newblock \href {https://doi.org/10.1126/science.1213847}
  {\path{doi:10.1126/science.1213847}}.

\bibitem{collaborationReproducibilityProjectModel2013}
O.~S. Collaboration, The {{Reproducibility Project}}: {{A Model}} of
  {{Large}}-{{Scale Collaboration}} for {{Empirical Research}} on
  {{Reproducibility}}, {{SSRN Scholarly Paper}} ID 2195999, {Social Science
  Research Network}, {Rochester, NY} (Jan. 2013).
\newblock \href {https://doi.org/10.2139/ssrn.2195999}
  {\path{doi:10.2139/ssrn.2195999}}.

\bibitem{ECJ19}
E.~O. Scott, S.~Luke, {ECJ} at 20: Toward a general metaheuristics toolkit, in:
  Proceedings of the Genetic and Evolutionary Computation Conference Companion,
  GECCO ’19, Association for Computing Machinery, New York, NY, USA, 2019, p.
  1391–1398.

\bibitem{wagner2014}
S.~Wagner, G.~Kronberger, A.~Beham, M.~Kommenda, A.~Scheibenpflug, E.~Pitzer,
  S.~Vonolfen, M.~Kofler, S.~Winkler, V.~Dorfer, M.~Affenzeller, Advanced
  Methods and Applications in Computational Intelligence, Vol.~6 of Topics in
  Intelligent Engineering and Informatics, Springer, 2014, Ch. Architecture and
  Design of the {HeuristicLab} Optimization Environment, pp. 197--261.

\bibitem{DN11}
J.~J. Durillo, A.~J. Nebro, {jMetal}: A java framework for multi-objective
  optimization, Advances in Engineering Software 42 (2011) 760--771.

\bibitem{NDV15}
A.~J. Nebro, J.~J. Durillo, M.~Vergne, Redesigning the {jMetal} multi-objective
  optimization framework, in: Proceedings of the Companion Publication of the
  2015 Annual Conference on Genetic and Evolutionary Computation, GECCO
  Companion ’15, Association for Computing Machinery, New York, NY, USA,
  2015, p. 1093–1100.

\bibitem{jmetalcpp13}
E.~L\'opez-Camacho, M.~J. Garc\'ia~Godoy, A.~J. Nebro, J.~F. Aldana-Montes,
  {jMetalCpp}: optimizing molecular docking problems with a c++ metaheuristic
  framework, Bioinformatics 30~(3) (2013) 437--438.

\bibitem{jMetalPy19}
A.~Ben\'itez-Hidalgo, A.~J. Nebro, J.~Garc\'ia-Nieto, I.~Oregi, J.~D. Ser,
  {jMetalPy}: A python framework for multi-objective optimization with
  metaheuristics, Swarm and Evolutionary Computation 51 (2019) 100598.

\bibitem{Hadka20}
D.~Hadka, \href{http://moeaframework.org/}{MOEA Framework. A Free and Open
  Source Java Framework for Multiobjective Optimization} (2020).
\newline\urlprefix\url{http://moeaframework.org/}

\bibitem{NiaPyJOSS2018}
G.~Vrban{\v{c}}i{\v{c}}, L.~Brezo{\v{c}}nik, U.~Mlakar, D.~Fister, I.~{Fister
  Jr.}, {NiaPy}: Python microframework for building nature-inspired algorithms,
  Journal of Open Source Software 3 (2018).

\bibitem{Pagmo20}
F.~Biscani, D.~Izzo, \href{https://esa.github.io/pagmo2/}{pagmo} (Jan. 2020).
\newline\urlprefix\url{https://esa.github.io/pagmo2/}

\bibitem{Paradiseo04}
S.~Cahon, N.~Melab, E.-G. Talbi, Paradiseo: A framework for the reusable design
  of parallel and distributed metaheuristics, Journal of Heuristics (2004).

\bibitem{PlatEMO17}
Y.~Tian, R.~Cheng, X.~Zhang, Y.~Jin, {PlatEMO}: A {MATLAB} platform for
  evolutionary multi-objective optimization, IEEE Computational Intelligence
  Magazine 12~(4) (2017) 73--87.

\bibitem{Pygmo20}
F.~Biscani, D.~Izzo, \href{https://esa.github.io/pygmo2/}{pygmo} (Jan. 2020).
\newline\urlprefix\url{https://esa.github.io/pygmo2/}

\bibitem{Platypus20}
D.~Hadka, \href{https://platypus.readthedocs.io/}{Platypus - Multiobjective
  Optimization in Python} (2020).
\newline\urlprefix\url{https://platypus.readthedocs.io/}

\bibitem{HLUY20}
C.~{Huang}, Y.~{Li}, X.~{Yao}, A survey of automatic parameter tuning methods
  for metaheuristics, IEEE Transactions on Evolutionary Computation 24~(2)
  (2020) 201--216.

\bibitem{LDP+16}
M.~L.-I. nez, J.~Dubois-Lacoste, L.~{P\'erez C\'aceres}, M.~Birattari,
  T.~St\"utzle",
  \href{http://www.sciencedirect.com/science/article/pii/S2214716015300270}{The
  irace package: Iterated racing for automatic algorithm configuration},
  Operations Research Perspectives 3 (2016) 43 -- 58.
\newblock \href {https://doi.org/https://doi.org/10.1016/j.orp.2016.09.002}
  {\path{doi:https://doi.org/10.1016/j.orp.2016.09.002}}.
\newline\urlprefix\url{http://www.sciencedirect.com/science/article/pii/S2214716015300270}

\bibitem{HHL+09}
F.~Hutter, H.~H. Hoos, K.~Leyton-Brown, T.~St\"{u}tzle, Paramils: An automatic
  algorithm configuration framework, J. Artif. Int. Res. 36~(1) (2009)
  267–306.

\bibitem{gabrel2014recent}
V.~Gabrel, C.~Murat, A.~Thiele, Recent advances in robust optimization: An
  overview, European journal of operational research 235~(3) (2014) 471--483.

\bibitem{jin2005evolutionary}
Y.~Jin, J.~Branke, Evolutionary optimization in uncertain environments-a
  survey, IEEE Transactions on evolutionary computation 9~(3) (2005) 303--317.

\bibitem{paenke2006efficient}
I.~Paenke, J.~Branke, Y.~Jin, Efficient search for robust solutions by means of
  evolutionary algorithms and fitness approximation, IEEE Transactions on
  Evolutionary Computation 10~(4) (2006) 405--420.

\bibitem{ben1999robust}
A.~Ben-Tal, A.~Nemirovski, Robust solutions of uncertain linear programs,
  Operations research letters 25~(1) (1999) 1--13.

\bibitem{jin2003trade}
Y.~Jin, B.~Sendhoff, Trade-off between performance and robustness: An
  evolutionary multiobjective approach, in: International Conference on
  Evolutionary Multi-Criterion Optimization, Springer, 2003, pp. 237--251.

\bibitem{deb2009reliability}
K.~Deb, S.~Gupta, D.~Daum, J.~Branke, A.~K. Mall, D.~Padmanabhan,
  Reliability-based optimization using evolutionary algorithms, IEEE
  Transactions on Evolutionary Computation 13~(5) (2009) 1054--1074.

\bibitem{van2020towards}
K.~van~der Blom, T.~M. Deist, T.~Tu{\v{s}}ar, M.~Marchi, Y.~Nojima, A.~Oyama,
  V.~Volz, B.~Naujoks, Towards realistic optimization benchmarks: A
  questionnaire on the properties of real-world problems, arXiv preprint
  arXiv:2004.06395 (2020).

\bibitem{dunning2017jump}
I.~Dunning, J.~Huchette, M.~Lubin, Jump: A modeling language for mathematical
  optimization, SIAM Review 59~(2) (2017) 295--320.

\bibitem{noel2012new}
M.~M. Noel, A new gradient based particle swarm optimization algorithm for
  accurate computation of global minimum, Applied Soft Computing 12~(1) (2012)
  353--359.

\bibitem{bonissone2006evolutionary}
P.~P. Bonissone, R.~Subbu, N.~Eklund, T.~R. Kiehl, Evolutionary algorithms+
  domain knowledge= real-world evolutionary computation, IEEE Transactions on
  Evolutionary Computation 10~(3) (2006) 256--280.

\bibitem{fischetti2018matheuristics}
M.~Fischetti, M.~Fischetti, Matheuristics, in: Handbook of Heuristics,
  Springer, 2018, pp. 121--153.

\bibitem{wu2015variable}
G.~Wu, W.~Pedrycz, P.~N. Suganthan, R.~Mallipeddi, A variable reduction
  strategy for evolutionary algorithms handling equality constraints, Applied
  Soft Computing 37 (2015) 774--786.

\bibitem{das2010problem}
S.~Das, P.~N. Suganthan, Problem definitions and evaluation criteria for cec
  2011 competition on testing evolutionary algorithms on real world
  optimization problems, Jadavpur University, Nanyang Technological University,
  Kolkata (2010) 341--359.

\bibitem{juan2015review}
A.~A. Juan, J.~Faulin, S.~E. Grasman, M.~Rabe, G.~Figueira, A review of
  simheuristics: Extending metaheuristics to deal with stochastic combinatorial
  optimization problems, Operations Research Perspectives 2 (2015) 62--72.

\bibitem{chica2017simheuristics}
M.~Chica, J.~P{\'e}rez, A.~Angel, O.~Cordon, D.~Kelton, Why simheuristics?
  benefits, limitations, and best practices when combining metaheuristics with
  simulation, Benefits, Limitations, and Best Practices When Combining
  Metaheuristics with Simulation (January 1, 2017) (2017).

\bibitem{jin2011surrogate}
Y.~Jin, Surrogate-assisted evolutionary computation: Recent advances and future
  challenges, Swarm and Evolutionary Computation 1~(2) (2011) 61--70.

\bibitem{jin2005comprehensive}
Y.~Jin, A comprehensive survey of fitness approximation in evolutionary
  computation, Soft computing 9~(1) (2005) 3--12.

\bibitem{rasheed2000informed}
K.~Rasheed, H.~Hirsh, Informed operators: Speeding up genetic-algorithm-based
  design optimization using reduced models, in: Proceedings of the 2nd Annual
  Conference on Genetic and Evolutionary Computation, 2000, pp. 628--635.

\bibitem{jin2002framework}
Y.~Jin, M.~Olhofer, B.~Sendhoff, A framework for evolutionary optimization with
  approximate fitness functions, IEEE Transactions on evolutionary computation
  6~(5) (2002) 481--494.

\bibitem{bhosekar2018advances}
A.~Bhosekar, M.~Ierapetritou, Advances in surrogate based modeling, feasibility
  analysis, and optimization: A review, Computers \& Chemical Engineering 108
  (2018) 250--267.

\bibitem{arrieta2020explainable}
A.~B. Arrieta, N.~D{\'\i}az-Rodr{\'\i}guez, J.~Del~Ser, A.~Bennetot, S.~Tabik,
  A.~Barbado, S.~Garc{\'\i}a, S.~Gil-L{\'o}pez, D.~Molina, R.~Benjamins,
  R.~Chatila, F.~Herrera, Explainable artificial intelligence (xai): Concepts,
  taxonomies, opportunities and challenges toward responsible ai, Information
  Fusion 58 (2020) 82--115.

\bibitem{guo2018survey}
R.~Guo, L.~Cheng, J.~Li, P.~R. Hahn, H.~Liu, A survey of learning causality
  with data: Problems and methods, arXiv preprint arXiv:1809.09337 (2018).

\bibitem{moraffah2020causal}
R.~Moraffah, M.~Karami, R.~Guo, A.~Raglin, H.~Liu, Causal interpretability for
  machine learning-problems, methods and evaluation, ACM SIGKDD Explorations
  Newsletter 22~(1) (2020) 18--33.

\bibitem{huang2020survey}
C.~Huang, Y.~Li, X.~Yao, A survey of automatic parameter tuning methods for
  metaheuristics, IEEE Transactions on Evolutionary Computation 24~(2) (2020)
  201--216.

\bibitem{smith2008towards}
K.~A. Smith-Miles, Towards insightful algorithm selection for optimisation
  using meta-learning concepts, in: 2008 IEEE International Joint Conference on
  Neural Networks (IEEE World Congress on Computational Intelligence), ieee,
  2008, pp. 4118--4124.

\bibitem{kotthoff2016algorithm}
L.~Kotthoff, Algorithm selection for combinatorial search problems: A survey,
  in: Data Mining and Constraint Programming, Springer, 2016, pp. 149--190.

\bibitem{smith2011discovering}
K.~Smith-Miles, J.~van Hemert, Discovering the suitability of optimisation
  algorithms by learning from evolved instances, Annals of Mathematics and
  Artificial Intelligence 61~(2) (2011) 87--104.

\bibitem{kanda2016meta}
J.~Kanda, A.~de~Carvalho, E.~Hruschka, C.~Soares, P.~Brazdil, Meta-learning to
  select the best meta-heuristic for the traveling salesman problem: A
  comparison of meta-features, Neurocomputing 205 (2016) 393--406.

\bibitem{gutierrez2019selecting}
A.~E. Gutierrez-Rodr{\'\i}guez, S.~E. Conant-Pablos, J.~C. Ortiz-Bayliss,
  H.~Terashima-Mar{\'\i}n, Selecting meta-heuristics for solving vehicle
  routing problems with time windows via meta-learning, Expert Systems with
  Applications 118 (2019) 470--481.

\bibitem{pavelski2019meta}
L.~M. Pavelski, M.~R. Delgado, M.-{\'E}. Kessaci, Meta-learning on flowshop
  using fitness landscape analysis, in: Proceedings of the Genetic and
  Evolutionary Computation Conference, 2019, pp. 925--933.

\bibitem{wu2019ensemble}
G.~Wu, R.~Mallipeddi, P.~N. Suganthan, Ensemble strategies for population-based
  optimization algorithms--a survey, Swarm and evolutionary computation 44
  (2019) 695--711.

\end{thebibliography}

\end{document}